%% file: main.tex
\documentclass{article}
\usepackage{amsbsy,mathtools,multirow,cases}
\usepackage{tabularx}
\usepackage{fullpage}
\usepackage{pgfplots}
\usepackage{subcaption}

\usepackage{arxiv}
\usepackage{graphicx}%
\usepackage{multirow}%
\usepackage{amsmath,amssymb,amsfonts}%
\usepackage{mathrsfs}%
\usepackage[title]{appendix}%
\usepackage{xcolor}%
\usepackage{textcomp}%
\usepackage{manyfoot}%
\usepackage{booktabs}%
\usepackage{algorithm}%
\usepackage{algorithmicx}%
\usepackage{algpseudocode}%
\usepackage{listings}%
\usepackage{hyperref}
\usepackage{xr}

\title{CoCoGen: Physically-Consistent and Conditioned Score-based Generative Models for Forward and Inverse Problems}

\author{{Christian Jacobsen}\\
        Department of Aerospace Engineering\\
        University of Michigan\\ 
        Ann Arbor, MI 48109\\
        \texttt{csjacobs@umich.edu}\\
        \And
        {Yilin Zhuang}\\
        Department of Aerospace Engineering\\
        University of Michigan\\
        Ann Arbor, MI 48109\\
        \texttt{ylzhuang@umich.edu}\\
        \And
        {Karthik Duraisamy} \\
        Department of Aerospace Engineering \\
        Michigan Institute for Computational Discovery \& Engineering \\
        University of Michigan \\
        Ann Arbor, MI 48109 \\
        \texttt{kdur@umich.edu} \\
        }

\begin{document}


    


\maketitle
\begin{abstract}

\input{00_abstract}
\end{abstract}





\input{01_introduction}
\input{02_background}
\input{03_physical_consistency}

\input{04_unconditional}
\input{05_conditional}

\input{06_conclusion}

\bibliographystyle{elsarticle-num}
\bibliography{references}

 \begin{appendices}
     \input{appendix}
 \end{appendices}

\end{document}

%% file: 00_abstract.tex
Recent advances in generative artificial intelligence have had a significant impact on diverse domains spanning computer vision, natural language processing, and drug discovery. This work extends the reach of generative models into physical problem domains, particularly addressing the efficient enforcement of physical laws  and conditioning for forward and inverse problems involving partial differential equations (PDEs). 
Our work introduces two key contributions: firstly, we present an efficient approach to promote consistency with the underlying PDE. By incorporating discretized information into score-based generative models, our method generates samples closely aligned with the true data distribution, showcasing residuals comparable to data generated through conventional PDE solvers, significantly enhancing fidelity. Secondly, we showcase the potential and versatility of score-based generative models in various physics tasks, specifically highlighting surrogate modeling as well as probabilistic field reconstruction and inversion from sparse measurements. A robust foundation is laid by designing unconditional score-based generative models that utilize  reversible probability flow ordinary differential equations. Leveraging  conditional models that require minimal training, we illustrate their flexibility when combined with a frozen unconditional model. These conditional models generate PDE solutions by incorporating  parameters, macroscopic quantities, or partial field measurements as guidance. The results illustrate the inherent flexibility of score-based generative models and explore the synergy between unconditional score-based generative models and the present physically-consistent sampling approach, emphasizing the power and flexibility in solving for and inverting physical fields governed by differential equations, and in other scientific machine learning tasks. 

%% file: 01_introduction.tex
\section{Introduction}\label{sec:intro}

Generative modeling techniques aim to sample from an unknown data distribution given only samples from the distribution. These methods can be largely classified into three main categories: likelihood-based models, implicit generative models, and the more recent diffusion and score-based generative models. A large array of likelihood-based techniques exist in the literature such as variational autoencoders (VAEs)~\cite{kingma2014autoencoding,pmlr-v32-rezende14, Jacobsen2022}, autoregressive models~\cite{pmlr-v15-larochelle11a,pmlr-v37-germain15,pmlr-v48-oord16}, energy-based models (EBMs)~\cite{ebm1,song2021train}, and normalizing flow models~\cite{pmlr-v37-rezende15,dinh2015nice,dinh2017density}. Likelihood-based approaches have been shown to be useful, yet impose restrictions on the model or distribution form to ensure a regularized distribution and facilitate likelihood computation during training. Implicit generative models such as generative adversarial networks (GANs)~\cite{NIPS2014_5ca3e9b1} are challenging to train due to the adversarial nature of training. Score-based approaches, on the other hand, model the score function (gradient of the log density) directly, bypassing restrictions on model or distribution form to represent a valid probability density~\cite{pmlr-v48-chwialkowski16}. 
This leads to much greater flexibility in allowable model architectures and facilitates the learning of more complex distributions over other methods. Although many such models are often referred to as \emph{diffusion models} in the literature, we mostly refer to them as \emph{score-based models} which contain diffusion models as a subset. 

The pioneering work of Song et al.~\cite{song2021scorebased} ties the earliest diffusion models together under a common framework of score-based modeling, a framework which our work is based on. Some of the earliest score-based generative models include de-noising diffusion probabilistic models (DDPM)~\cite{NEURIPS2020_4c5bcfec} and de-noising score matching with Langevin dynamics (SMLD)~\cite{NEURIPS2019_3001ef25}. These early models initially showed great potential in advancing the state of the art in generative modeling. However, more recent approaches such as score-based generative modeling with SDEs~\cite{song2021scorebased}, Poisson flow generative models (PFGM)~\cite{xu2022poisson,xu2023pfgm}, and consistency trajectory models (CTM)~\cite{kim2023consistency} suggest that score-based generative models may be a superior choice among generative modeling techniques.

Diffusion models and score-based generative models have demonstrated state-of-the-art results in a variety of fields including natural language processing~\cite{austin2021structured, li2022diffusionlm, gong2023diffuseq, osti_10433150}, computer vision~\cite{9887996, repaint, 10.1145/3528233.3530757, 10203350},  multi-model learning~\cite{rombach2022high, ramesh2022hierarchical, nichol2022glide, ruiz2022dreambooth, controlnet}, drug design~\cite{guan2023d, luo2022antigenspecific}, and medical imaging~\cite{song2022solving, CHUNG2022102479}, also opening new avenues of exploration in their respective domains. 
Models based on diffusion and score-based generation overwhelmingly dominate the current state-of-the-art~\cite{kim2023consistency} in generative tasks such as the Frechet inception distance (FID) score on the popular CIFAR-10 dataset~\cite{Krizhevsky2009LearningML}. We therefore do not address comparisons to other classes of generative models in this work, but rather focus on a new breadth of areas to advance and apply these models to. A comprehensive study of the available methods and applications of diffusion models and score-based generative models is included in Ref.~\cite{10.1145/3626235}. 

Diffusion and score-based models present great potential to be applied in areas of physics involving data augmentation, surrogate modeling, super-resolution, reduced order modelling, field inversion, field reconstruction, and uncertainty quantification among others. To the best of our knowledge, however, previous works have not thoroughly addressed enforcing adherence to physical laws. Some recent works have applied diffusion or score-based models to problems involving fluid flows, but little work on general problems governed by partial differential equations (PDEs) currently exists. 
Yang et al.~\cite{yang2023denoising} develop a surrogate model named FluidDiff based on diffusion models is developed to predict flows governed by the 2D incompressible Navier-Stokes equations given a source function. The model is shown to achieve greater accuracy in prediction over other methods such as conditional GANs~\cite{NIPS2014_5ca3e9b1}, U-Nets~\cite{10.1007/978-3-319-24574-4_28}, and even physics-informed neural networks~\cite{RAISSI2019686} (PINNs) in some cases. The primary metric of comparison is the root mean-squared error (RMSE) without emphasis on the adherence of predictions to the physical equations. Ultimately, without enforcement of the physical equations, the model may inevitably predict non-physical behavior. 
Another important work develops an iterative method for performing field reconstruction from sparse measurements on 2D flows governed by the Kolmogorov flow equations~\cite{SHU2023111972}. The model is based on conditional DDPMs in which the conditioning information is a low resolution guess of the full field along with gradients of the physical residual. Accurate high-fidelity reconstructions are observed along with reasonably good adherence to the physical equations of interest; however, there is no enforcement of physical behavior. Residual gradients are included as conditioning information, but there is no enforcement towards minimizing the residual during sampling. This in turn also allows for the model to potentially predict non-physical behavior.

Our work thus aims to address the issue of enforcing physical consistency in score-based generative models. We do not claim to develop state-of-the-art models for performing the tasks described here, but rather demonstrate the effectiveness and flexibility that score-based models present in addressing a wide range of physics-based problems.   Physical consistency enforcement involves ensuring or encouraging model predictions to follow physical equations of interest.  We develop a novel and flexible method of enforcing score-based generative models to generate samples which are characterized by low residual with respect to a nonlinear PDE describing some physical system. This method is different from PINN-type methods~\cite{RAISSI2019686} in many ways: i) The goal is to not solve a PDE, but rather to generate samples constrained by them; ii) Rather than using the PDE residual during the training stage, the residual is utilized during the sample generation stage; iii) Rather than relying on the continuous form of the residual, a discrete form is utilized based on the defined data. This method of sampling along with other aspects of the sampling process are fully scrutinized and illustrated to greatly reduce residuals of generated samples over the baseline sampling method on an unconditional model. We further exemplify the power of physical consistency enforcement by illustrating it in tandem with two applications. 
At this juncture, we remark that the work by Chung et al.~\cite{CHUNG2022102479} is most closely related to some aspects of the work presented here, although applied to image reconstruction. Corrector steps are applied during sampling from the score-based model to encourage sample consistency with a linear system modeling the data. Our work similarly contains correction steps; however, the nature of the correction steps is quite different and achieved in a different manner. The `correction steps' in our work constitute minimizing the residual from general nonlinear physical PDEs rather than a linear system to perform image reconstruction.

Conditional score-based models based on a frozen unconditional backbone are created for each of the applications, adapting the ideas of ControlNet~\cite{controlnet} to create flexible conditional augmentations. Both applications are illustrated on 2D data governed by Darcy's law describing flow through porous media. We first train a conditional score-based generative model as a surrogate model for predicting fluid flows as a function of the underlying parameters which determine the entire fluid flow. A second application performing probabilistic field reconstruction and inversion of Darcy flow fields is achieved by similarly training a conditional augmentation to the unconditional model. Pressure and permeability fields are predicted from sparsely sampled pressure measurements as the conditioning information. Sampling from the conditional model given such measurements consistently results in generated samples which adhere to the physical equations of interest with accurate reconstruction.

The organization of this manuscript is as follows: Section~\ref{sec:background} is a self-contained introduction to score-based generative models with SDEs, largely based on the work of Song et al~\cite{song2021scorebased}. Section~\ref{sec:physical_consistency} details our particular selection of the forward and reverse SDEs along with details of the training strategy in the unconditional and conditional settings. Our novel sampling approach which enforces physical consistency by incorporating the PDEs into the sampling process is also described in depth. Section~\ref{sec:unconditional} illustrates the use of physical consistency steps to ensure samples adhere to the equations of interest in an unconditional setting. Section~\ref{sec:conditional} presents  conditional generation in combination with our sampling approach to guide sample generation. Applications are illustrated in this conditional setting and include full-field generation and surrogate modeling, and probabilistic field reconstruction and inversion from sparse measurements. We provide our concluding remarks and directions for future work in Sec.~\ref{sec:conclusion}. 


\begin{figure}[h!]
    \centering
    \includegraphics[width=\linewidth]{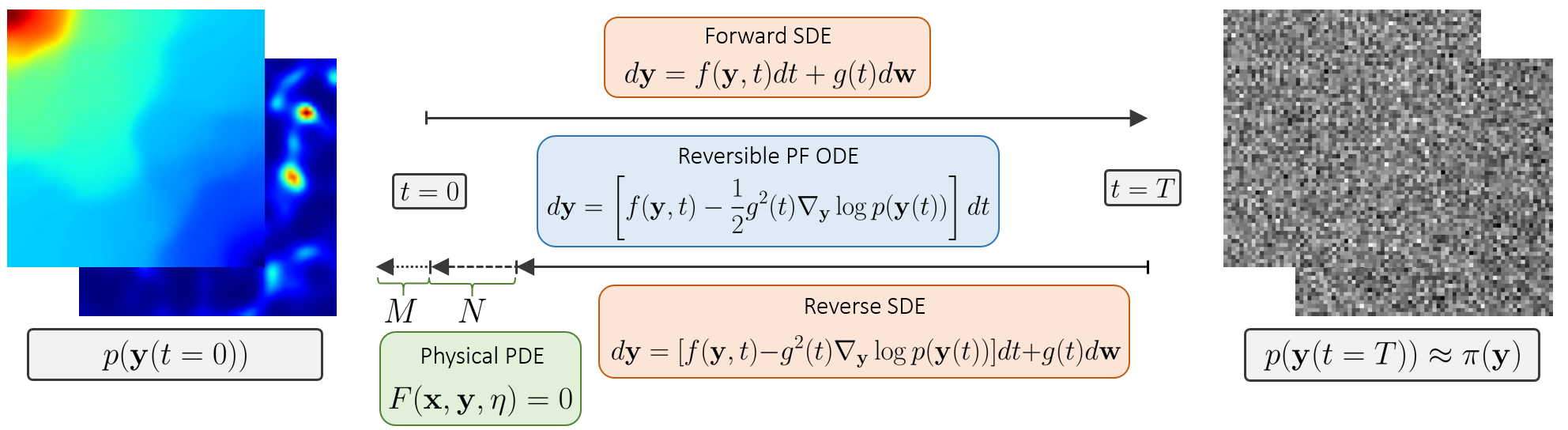}
    \caption{We propose a method of injecting the governing equations into the sampling process of score-based generative models to enforce consistency of samples with the underlying PDE.}
    \label{fig:main_fig}
\end{figure}

%% file: 02_background.tex
\section{Background}\label{sec:background}

The overarching goal of generative modelling is to sample from an unknown data distribution given only samples in the form of data from such a distribution. Unconditional generative modelling aims to sample from $p(\mathbf{y})$ given $N$ independent identically distributed (i.i.d.) samples $\{\mathbf{y}^{(i)}\}_{i=1}^N$ from the distribution. However, often there exist additional quantities in the form of conditional variables $\theta$ which contain information about data samples $\mathbf{y}$. In this case, a conditional generative model can be sought to enable sampling from the conditional distribution $p(\mathbf{y}|\theta)$. We explore both unconditional (Sec.~\ref{sec:physical_consistency}) and conditional (Sec.~\ref{sec:conditional}) score-based generative models in the context of forward and inverse problems,  beginning by describing the unconditional form. 

Score-based generative models are a class of generative modeling technique in which data is used to learn the Stein score function~\cite{pmlr-v48-chwialkowski16} (gradient of the log-density $\nabla_\mathbf{y} \log p(\mathbf{y})$). Once the score function is approximated, sampling from the distribution $p(\mathbf{y})$ can be performed using an iterative process known as Langevin dynamics~\cite{PARISI1981378,400cc48a-c98e-3a92-ac96-5477dc6f3a71}. However, in practice it is difficult to accurately estimate the score function in all regions of the data space due to data sparsity. Recent works~\cite{NEURIPS2019_3001ef25,NEURIPS2020_4c5bcfec} have thus suggested a key process of iteratively adding varying levels of noise to data samples and estimating the score function at each noise level. Adding noise effectively changes the underlying distribution, reducing sparsity in the data space and allowing for accurate score estimation in broader regions of the data space when compared to the original data distribution. Noise is sequentially added until the data is approximately distributed according to a prior distribution which is easy to sample from. Sample generation is then performed as the reverse of this process: sampling from the prior distribution and iteratively removing the noise until the data sample approximately belongs to the original data distribution. We refer readers to Ref.~\cite{songblog} for an intuitive and thorough explanation of score-based generative models. 



This iterative process of sequentially adding noise to data can be viewed in a continuous sense as the solution to a stochastic differential equation (SDE)~\cite{song2021scorebased}. We consider a data distribution at $t=0$ given by $p(\mathbf{y}(t=0))$ where $t$ does not represent physical time, but is rather a modelling quantity used to simulate the SDE. This is the distribution from which data samples $\{\mathbf{y}^{(i)}\}_{i=1}^N$ are drawn, where $\mathbf{y}^{(i)} \sim p(\mathbf{y}(t=0))$ and $\mathbf{y} \in \mathcal{X}$ (the data space). A stochastic differential equation (SDE) of the form 
\begin{equation}\label{eq:sde}
    d\mathbf{y} = \mathbf{f}(\mathbf{y}, t)dt + g(t)d\mathbf{w}
\end{equation}
describes the `dynamics' of adding noise to data samples, where $\mathbf{f}:\mathbb{R}^n\times \mathbb{R} \rightarrow \mathbb{R}^n$ is a function known as the \textit{drift coefficient}, $g:\mathbb{R}\rightarrow \mathbb{R}$ is a function known as the \textit{diffusion coefficient}, and $\mathbf{w}$ denotes standard Brownian motion. The system `dynamics' are constructed such that the distribution $p(\mathbf{y}(t=T))$ at final time $T$ approximately follows a prior distribution $\pi(\mathbf{y})$ which is easy to sample from, such as the standard normal distribution. The particular form of $\mathbf{f}(\mathbf{y}, t)$ and $g(t)$ are chosen such that $p(\mathbf{y}(t=T))\approx \pi(\mathbf{y})$, or vice-versa. The SDE describes dynamics of adding random noise to samples from $p(\mathbf{y}(t=0))$ until $\pi(\mathbf{y})$ is achieved. Suppose a sample $\mathbf{y}(t=0)$ is drawn from  $p(\mathbf{y}(t=0))$, and the dynamics of the SDE are simulated. Viewing $\mathbf{y}(t)$ at any intermediate time step $t \in (0, T]$ is essentially a noisy version of the initial sample. In this work, we assume a final time $T=1$ in all experiments, though this is simply a convention. 

In this continuous setting, the noising process can be reversed by using a well-known result~\cite{ANDERSON1982313} to solve the \emph{reverse SDE} backward in time. This reverse SDE is given by
\begin{equation}\label{eq:sde_reverse}
    d\mathbf{y} = [\mathbf{f}(\mathbf{y},t)-g^2(t)\nabla_\mathbf{y} \log p(\mathbf{y}(t))]dt + g(t)d \mathbf{w}\; ,
\end{equation}
where the distribution at any time $p(\mathbf{y}(t))$ is identical when solving the forward or reverse SDE. 
As $\mathbf{f}$ and $g$ are chosen particularly such that $p(\mathbf{y}(t=T)) \approx \pi(\mathbf{y})$, a sample $\mathbf{y}(t=T)\sim \pi(\mathbf{y})$ is drawn and the reverse SDE (Eq.~\ref{eq:sde_reverse}) is simulated backward in time to obtain a sample $\mathbf{y}(t=0)$ which is approximately drawn from $p(\mathbf{y}(t=0))$. 

To solve the reverse SDE, the score function $\nabla_\mathbf{y} \log p(\mathbf{y}(t))$ must be known for all times $t \in [0, T]$. Therefore, score-based generative modeling aims to model the score function using a parameterized model such as CNN-based or transformer-based machine learning models. This model $s_\phi(\mathbf{y}(t), t)$ should approximate the score function for all times $t\in[0, T]$ such that $s_\phi(\mathbf{y}(t), t)\approx \nabla_\mathbf{y} \log p(\mathbf{y}(t))$ and the dynamics of Eq.~\ref{eq:sde_reverse} are approximated by 
\begin{equation}\label{eq:sde_reverse_approx}
    d\mathbf{y} = [\mathbf{f}(\mathbf{y},t)-g^2(t)s_\phi(\mathbf{y}(t), t)]dt + g(t)d\mathbf{w}\; .
\end{equation}

The functions $\mathbf{f}$ and $g$ are often selected such that they induce a Gaussian transition kernel
\begin{equation}\label{eq:transition_kernel}
    p(\mathbf{y}(t)|\mathbf{y}(0)) = \mathcal{N}(\mathbf{y}(t); \boldsymbol\mu(\mathbf{y}(0), t), \boldsymbol\sigma^2(\mathbf{y}(0), t)I) \; ,
\end{equation}
where $\mathbf{y}(0)$ denotes a sample $\mathbf{y}(t=0)$ from the data distribution. This means that the score of the transition kernel can be computed for any time $t$ without solving the forward SDE as $\nabla_\mathbf{y}\log p(\mathbf{y}(t)|\mathbf{y}(0)) = (\mu(\mathbf{y}(0), t) - \mathbf{y}(t))(\mu(\mathbf{y}(0), t)-\mathbf{y}(t))^T/\sigma^2(\mathbf{y}(0), t)$.

The model $s_\phi(\mathbf{y}(t), t)$ is trained using a score-matching objective~\cite{song2021scorebased}. In our work, we use denoising score-matching (DSM)~\cite{6795935, song2021scorebased} in which the loss function is given by
\begin{equation}\label{eq:score_matching}
    \min_\phi \mathbb{E}_{t\sim\mathcal{U}[0,T]} \left [\lambda(t)\mathbb{E}_{p(\mathbf{y}(0))p(\mathbf{y}(t)|\mathbf{y}(0))}[||s_\phi(\mathbf{y}(t), t) - \nabla_{\mathbf{y}(t)}\log p(\mathbf{y}(t)|\mathbf{y}(0))||_2^2] \right ]
\end{equation}
where $\lambda(t) : [0, 1] \rightarrow \mathbb{R}_{>0}$ is a positive weighting function and $p(t)\sim \mathcal{U}[0, T]$. A closed form solution for the quantity $\nabla_{\mathbf{y}(t)}\log p(\mathbf{y}(t)|\mathbf{y}(0))$ can be easily obtained from the transition kernel in Eq.~\ref{eq:transition_kernel}. We note that when $\lambda(t)=1/2$, the optimal solution to Eq.~\ref{eq:score_matching} corresponds to $s_\phi(\mathbf{y}(t), t)=\nabla_{\mathbf{y}}\log p(\mathbf{y}(t))$ almost surely~\cite{6795935}, which is sufficient to solve the reverse SDE in Eq.~\ref{eq:sde_reverse}. 

The probability flow ordinary differential equation (PF ODE) is similar to the reverse SDE in that it can be solved in reverse time to approximately sample from $p(\mathbf{y}(t=0))$ given the distribution at $p(\mathbf{y}(t=T))$. Although this equation is an ODE rather than an SDE, the marginals at each intermediate time $t$ of the reverse SDE and PF ODE are identical~\cite{song2021scorebased}. The PF ODE is reversible in time and given by
\begin{equation}\label{eq:probability_flow_ode}
    d\mathbf{y} = \left [ \mathbf{f}(\mathbf{y},t)-\frac{1}{2}g^2(t)\nabla_x \log p(\mathbf{y}(t)) \right ]dt \; .
\end{equation}
As illustrated in Sec.~\ref{sec:unconditional}, solving the PF ODE offers advantages in terms of accuracy in physics-based applications. This is likely due to the absence of a noise injection term, which may allow for better denoising capabilities during sampling. However, in some cases such as the RePaint~\cite{repaint} algorithm exemplified in Sec.~\ref{app:firms} of the provided supplementary materials, the reverse SDE offers superior performance in terms of efficiency. We will discuss the use of  both of these equations to sample from trained models throughout this work. 

%% file: 03_physical_consistency.tex
\section{Enforcing Physical Consistency and Conditioning}\label{sec:physical_consistency}
Score-based generative models have seen great success in generating natural images and replicating human-like creativity in some tasks. However, generating solutions to physical problems requires more quantitative rigor to ensure that the generated samples are consistent with the particular physical laws of interest. Our aim is to develop a procedure for sampling from score-based generative models which will ensure that generated samples adhere to some specified governing physical equations. If all samples generated satisfy the equations of interest, this unlocks many potential applications for score-based generative models and prevents such models from predicting non-physical behaviors. 

We consider general steady-state partial differential equations (PDEs) of the form 
\begin{equation}
    F(\mathbf{x}, \mathbf{y}, \mathbf{\eta}) = 0 \; ,
\end{equation}
where $\mathbf{x}$ are the spatial coordinates, $\mathbf{y}$ are the physical variables of interest, and $\eta$ are parameters describing the physical system. In our examples, solutions $\mathbf{y}$ are computed on a discretized spatial domain such that a single solution -- or data sample -- is given on a 2D physical domain $\mathcal{X} = \mathbb{R}^{n\times n}$. We remark that constraining samples to follow the governing PDE is done {\bf only at inference/sample time, and does not have any effect on the training procedure.} This allows for much greater flexibility without the need for re-training the model. We first discuss training a score-based generative model before describing our approach to enforcing physical consistency in generated samples. 

\subsection{Unconditional Model Training}

\begin{figure}[h!]
    \centering
    \includegraphics[width=0.8\linewidth]{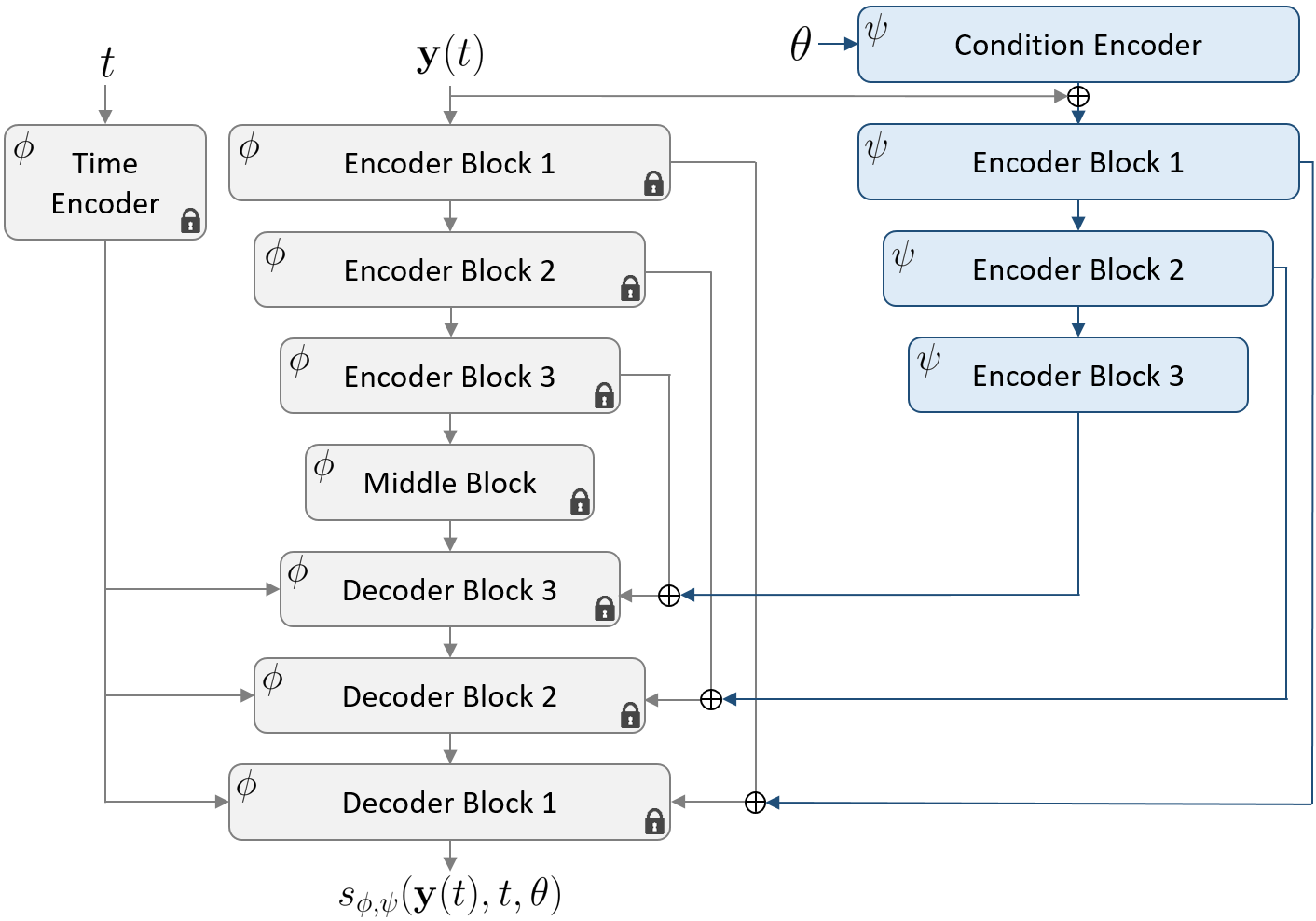}
    \caption{ControlNet-type architecture based on a UNet-type unconditional model architecture. The conditional model contains a fixed pretrained unconditional model with parameters $\phi$ and a conditional model augmentation with trainable parameters $\psi$.}
    \label{fig:controlnet_arch}
\end{figure}

We train a convolution-based UNet-type~\cite{10.1007/978-3-319-24574-4_28} architecture to approximate the score function in time with the weighting function set to a constant $\lambda(t) = 1$. The overall architecture for both our unconditional and conditional generative models is illustrated in Fig.~\ref{fig:controlnet_arch}. Our choice of $\mathbf{f}(\mathbf{y},t)$ and $g(t)$ in the SDE form correspond to the variance-preserving (VP) SDE~\cite{song2021scorebased} in which
\begin{align}
   \mathbf{f}(\mathbf{y}, t) & = -\frac{1}{2}\beta(t)\mathbf{y} \label{eq:VP_f}, \; \mathrm{and} \\
   g(t) & = \sqrt{\beta(t)} \label{eq:VP_g}  \; ,
\end{align}
giving a final forward SDE form of
\begin{equation}\label{eq:VP_forward_SDE}
    d\mathbf{y} = -\frac{1}{2}\beta(t)\mathbf{y} dt + \sqrt{\beta(t)} dw\; .
\end{equation}
Further, we define $\beta(t)$ as a linear function with two hyperparameters: 
\begin{equation}\label{eq:beta_linear}
    \beta(t) = \beta_{min} + (\beta_{max}-\beta_{min})t \; ,
\end{equation}
where $\beta_{min}, \beta_{max} > 0$ and $\beta_{max} > \beta_{min}$. 
This SDE form induces a transition kernel of
\begin{equation}\label{eq:VP_transition_kernel}
    p(\mathbf{y}(t)|\mathbf{y}(0)) = \mathcal{N}\left ( \mu(\mathbf{y}, t), \Sigma(t) \right )\; ,
\end{equation}
where
\begin{align}
    \mu(\mathbf{y},t) & = \mathbf{y}(0)\exp{\left [-\frac{1}{4}t^2(\beta_{max}-\beta_{min}) - \frac{1}{2}t\beta_{min} \right ]} \\
    \Sigma(t) & = \mathbf{I}\left (1-\exp{\left [ -\frac{1}{2}t^2(\beta_{max}-\beta_{min}) - t\beta_{min}\right ]}\right )
\end{align}
    
With a Gaussian transition kernel, the score function can be computed analytically such that we train with the following loss function:
\begin{equation}\label{eq:training_loss}
    \min_\theta \mathbb{E}_{t\sim \mathcal{U}[0,T]} \left [\mathbb{E}_{p(\mathbf{y}(0))p(\mathbf{y}(t)|\mathbf{y}(0))}[||s_\theta(\mathbf{y}(t), t) + z\Sigma^{1/2}(t)||_2^2] \right ] \; ,
\end{equation}
where $z \sim \mathcal{N}(0, I)$. During training, we set $\beta_{min}=1\times 10^{-4}$, $\beta_{max} = 10$, and $T=1$ for all of our experiments. 

\subsection{Learning Conditional Score Functions}\label{sec:controlnet}

Unconditional generators have diverse applications, such as sampling from $p(\mathbf{y})$ for data augmentation, distribution learning, and anomaly detection. Moreover, they can serve as a robust foundation for constructing conditional generative models. Such models aim to sample from $p(\mathbf{y}|\theta)$, where $\theta$ represents conditioning information, guiding the sampling process. Integrating conditioning into existing unconditional score-based generative models extends their utility, particularly in generating physical fields based on various input conditions like generative parameters, boundary conditions, partial field measurements, or macroscopic quantities. A conditional score-based generative model has the potential to undertake tasks like field reconstruction, field inversion, and effective probabilistic surrogate modeling. In some instances, an approximate conditional sampling can be achieved from a pre-trained unconditional model without additional training or conditional modeling. However, in most cases, generating samples from the desired conditional distribution requires extra data, modeling, and/or training efforts.

In scenarios where an analytical approximation to the conditional score function is unavailable, we resort to training a conditional model to approximate the conditional score function. Developing a conditional score-based generative model requires not only data samples $\mathbf{y}$ but also corresponding conditional information $\theta$ for each sample, forming pairs $\{\mathbf{y}^{(i)}, \theta^{(i)}\}_{i=1}^N$ in the dataset. The conditioning information could be a macroscopic quantity derived from the physical field, the generative parameters defining the physical field, or partial field measurements, among other possibilities. With these data samples, a conditional generative model aims to sample from the true conditional distribution $p(\mathbf{y}|\theta)$.

Since the forward Stochastic Differential Equation (SDE) noising process defined by Eq.~\ref{eq:VP_forward_SDE} is Markovian, it remains independent of conditioning, even when conditioning information is available. The forward process stays the same: $p(\mathbf{y}(t)|\mathbf{y}(0),\theta) = p(\mathbf{y}(t)|\mathbf{y}(0))$. However, the score approximation model $s_{\phi}(\mathbf{y}(t), t, \theta)$ is designed to accept conditioning information as an additional input. In the general conditional case, the score approximation model is trained by
\begin{equation}\label{eq:score_matching_conditional}
    \min_\phi \mathbb{E}_t \left [\lambda(t)\mathbb{E}_{p(\mathbf{y}(0),\theta)p(\mathbf{y}(t)|\mathbf{y}(0))}[||s_{\phi}(\mathbf{y}(t), t, \theta) - \nabla_{\mathbf{y}(t)}\log p(\mathbf{y}(t)|\mathbf{y}(0),\theta)||_2^2] \right ] \; ,
\end{equation}
effectively embedding conditioning information into the score approximation to learn $\nabla_\mathbf{y} \log p(\mathbf{y}(t)|\theta)$. 


Training a full-scale generative model is often quite expensive and can be difficult to properly tune with optimal hyperparameters, architectures, and training schedules. However, in some scenarios in which multiple downstream objectives are of interest from the same application, we aim to train a single unconditional model and augment the unconditional model with smaller conditional models which can be easily connected and disconnected to facilitate generation using multiple different types of conditioning. To achieve this, we turn towards recent advancements in conditional score-based generative modeling and adapt them for our applications. In particular, we leverage ControlNet~\cite{controlnet}, a form of model architecture specifically designed to augment pretrained unconditional models with conditional generative capabilities. The idea of ControlNet is to freeze the pretrained unconditional model and connect it to a smaller trainable model which will incorporate the conditional information. The overall model architecture including the unconditional model and conditional augmentation which we employ is shown in Fig.~\ref{fig:controlnet_arch}. The conditional portion of the model to be trained is initialized with special zero-convolution~\cite{controlnet} layers such that the model will produce unhindered unconditional samples at the onset of training. Thus, if a pretrained unconditional model is validated to generate physical fields which are consistent with the underlying physical PDE, the conditional model is already guaranteed to produce physically accurate solutions at the onset of training. 

During training, the unconditional model parameters are frozen. The conditional augmentation is trained using a very similar objective function to Eq.~\ref{eq:score_matching_conditional} with the main exception that the conditional score function approximation $s_{\phi,\psi}(\mathbf{y}(t), t, \theta)$ contains non-trainable parameters $\phi$ and trainable parameters $\psi$. The training objective we train conditional augmentations with therefore becomes 
\begin{equation}\label{eq:score_matching_controlnet}
    \min_\psi \mathbb{E}_t \left [\lambda(t)\mathbb{E}_{p(\mathbf{y}(0),\theta)p(\mathbf{y}(t)|\mathbf{y}(0))}[||s_{\phi,\psi}(\mathbf{y}(t), t, \theta) - \nabla_{\mathbf{y}(t)}\log p(\mathbf{y}(t)|\mathbf{y}(0), \theta)||_2^2] \right ] \; .
\end{equation}

\subsection{Sampling}\label{sec:physical_consistency_sampling}
Sampling from the trained score-based generative model is
achieved by solving the reverse SDE or PF ODE backward in time. 
In our work, the particular form of reverse SDE solved during sampling is given by
\begin{equation}\label{eq:VP_reverse_SDE}
    d\mathbf{y} = \left [-\frac{1}{2}\beta(t)\mathbf{y} - \beta(t)s_\theta(\mathbf{y}(t),t)\right ]dt + \sqrt{\beta(t)}dw \; ,
\end{equation}
and the PF ODE by
\begin{equation}\label{eq:VP_reverse_PFODE}
    d\mathbf{y} = \left [-\frac{1}{2}\beta(t)\mathbf{y} - \frac{1}{2}\beta(t)s_\theta(\mathbf{y}(t),t)\right ]dt \; .
\end{equation}
These equations can be easily constructed from Eqs.~\ref{eq:sde_reverse},~\ref{eq:probability_flow_ode},~\ref{eq:VP_forward_SDE}, and~\ref{eq:beta_linear}. We note that the unconditional score function approximation $s_\phi(\mathbf{y}(t), t)$ is replaced with the conditional approximation $s_{\phi,\psi}(\mathbf{y}(t), t, \theta)$ in Eqs.~\ref{eq:VP_reverse_SDE} and~\ref{eq:VP_reverse_PFODE} to achieve conditional sampling. 

We assume a prior distribution of $\pi(\mathbf{y}) = \mathcal{N}(0, \mathbf{I})$. The transition kernel in Eq.~\ref{eq:transition_kernel} approaches this prior at $t\rightarrow \infty$, but with large enough difference in $\beta_{max}-\beta_{min}$, the transition kernel at $t=T=1$ will be close to the selected prior. 
After sampling from the prior, this becomes the initial condition of the reverse SDE or reverse PF ODE. Solving either of Eqs.~\ref{eq:VP_reverse_SDE} or~\ref{eq:VP_reverse_PFODE} from $t=1$ to $t=0$ will result in a sample drawn from the original data distribution $p(\mathbf{y}(0))$, as long as the score function approximation is accurate. 
These equations can be solved using any discretized SDE or ODE solver of choice. For simplicity, we opt to solve the reverse SDE in Eq.~\ref{eq:VP_reverse_SDE} using the first-order accurate Euler-Maruyama (EM)~\cite{numericalmethodsSDEs} method, and the PF ODE in Eq.~\ref{eq:VP_reverse_PFODE} with the first-order accurate forward Euler (FE) scheme. Defining $\tau$ as the number of discrete time steps, this corresponds to a timestep of $\Delta t = T/\tau$. 

To enforce physical consistency, we turn towards modifying the sampling process after training by reducing the physical residual $r$ at each point in the computational domain, where
\begin{equation}
    \mathbf{r} = F(\mathbf{x},\mathbf{y},\eta) \; .
\end{equation}
The residual constitutes computing the PDE operator on a {\bf discretized} approximate solution to the PDE. In practice, this residual will be nonzero, indicating that the discretized physical field $\mathbf{y}$ does not perfectly satisfy the physical equations. Our goal is to generate samples from a score-based generative model which minimizes this residual. Generated samples should contain residuals which are similar to the residual obtained by solving the PDE using traditional means. 
To reduce the residual at the final time of the sampling process, we aim to minimize the following expression at each point in the computational domain:
\[
\min_{\mathbf{y}(t=0)} \| \mathbf{r} \|^2_2 \; .
\]

To this end, we propose appending a residual minimization step to the SDE or ODE solver for the last $N$ of $\tau$ time steps, aiming to minimize $\| \mathbf{r} \|_2^2$ by appending a small step in the negative gradient direction $\nabla_\mathbf{y} \| \mathbf{r} \|^2_2 = 2\mathbf{r}\nabla_\mathbf{y} \mathbf{r}$. This requires evaluating the full order operator of the PDE and gradient of the residual by updating the sample according to
\begin{equation}\label{eq:physical_consistency}
    \mathbf{y}_{i-1} = \textrm{Solver}(\mathbf{y}_{i}, t_i) - 2\epsilon\mathbf{r}\nabla_\mathbf{y} \mathbf{r} \; ,
\end{equation}
where $\epsilon$ is the residual step size hyperparameter and $\textrm{Solver}(\mathbf{y}_i, t_i)$ indicates a step using the particular solver (EM for reverse SDE or FE for PF ODE). The residual is computed at each point in the computational domain, and each value is updated via physical consistency steps. After the reverse SDE process is solved to $t=0$, we propose performing an additional $M$ physical consistency steps by iteratively updating the final sample according to
\[
\mathbf{y}_{i-1} = \mathbf{y}_{i} - 2\epsilon\mathbf{r}\nabla_\mathbf{y} \mathbf{r} \; .
\]

The sampling process is  flexible, with tunable hyperparameters $\epsilon$, $\tau$, $N$, and $M$, which can balance the need for efficiency over the degree to which samples follow the governing equations. An in-depth investigation into the effects of these hyperparameters is performed in Sec.~\ref{sec:darcy_uncond}. In our experiments, we find empirically that setting $\epsilon$ too large or too small results in unimproved sampling. We use $\epsilon = 2\times 10^{-4} / \max \nabla_\mathbf{y} \mathbf{r}$ during sampling in all of our experiments, though we suspect that this form may not be optimal for all PDEs and datasets.

However, the residual minimization step appended to the solver can be implemented for any governing PDE. 
As the training process is not altered, a trained generative model is expected to produce samples which are relatively close to satisfying the equations. Thus, with an accurately trained generative model, relatively few physical consistency steps will be required to greatly reduce residuals and bring the generated samples closer to adhering to the physics. We illustrate this process on an example from fluid dynamics involving flow through porous media.

%% file: 04_unconditional.tex
\section{Unconditional Generation of Darcy Flow Fields}\label{sec:unconditional}
Before addressing forward and inverse problems, we first demonstrate the capabilities of physical consistency sampling by training an unconditional model. Samples from the unconditional model are distributed approximately as the true data distribution. However, we find empirically that such samples will generally not satisfy the PDE. It is possible that developing more advanced architectures and finely tuning the training procedure may result in consistently small physical residuals, but we illustrate that the use of physical consistency sampling negates this need altogether. Our primary demonstrative example is that of Darcy flow. 

The Darcy flow equations relate fluid pressure to the permeability of a porous media through which it flows. A permeability field $K(\mathbf{x})$ describes the degree to which a fluid can pass through the media at each location $\mathbf{x}$ in the spatial domain while a source function $f_s(\mathbf{x})$ models physical locations in which a source or sink of fluid are entering or exiting the system. The pressure $p(\mathbf{x})$ and velocity fields $\mathbf{u}(\mathbf{x})$ of the fluid then satisfy the following equations according to Darcy's law:
\begin{align}\label{eq:darcy}
    \mathbf{u}(\mathbf{x}) & = -K(\mathbf{x})\nabla p(\mathbf{x}), \;\;\; \mathbf{x} \in \mathcal{X} \nonumber \\
    \nabla \cdot \mathbf{u}(\mathbf{x}) & = f_s(\mathbf{x}), \;\;\; \mathbf{x} \in \mathcal{X} \\
    \mathbf{u}(\mathbf{x}) \cdot \hat{\mathbf{n}}(\mathbf{x}) & = 0, \;\;\; \mathbf{x} \in \partial\mathcal{X} \nonumber \\
    \int_{\mathcal{X}} p(\mathbf{x})d\mathbf{x} & = 0. \nonumber
\end{align}

We generate training samples of Darcy flow fields by setting a constant source function given by Eq.~\ref{eq:darcy_source} and sampling a random permeability field $K(\mathbf{x})$.
\begin{equation}\label{eq:darcy_source}
f_s(\mathbf{x}) = 
\begin{cases}
    r, & \;\; \vert x_i - \frac{1}{2}w\vert \leq \frac{1}{2}w, \;\; i = 1,2 \\
    -r, & \;\; \vert x_i-1+\frac{1}{2}w\vert \leq \frac{1}{2}w, \;\; i = 1,2 \\
    0, & \;\; \textrm{otherwise}
\end{cases} \; \; \; , 
\end{equation}
We sample $K(\mathbf{x})$ by modeling the log-permeability field as a Gaussian random field with covariance function $k$
\begin{equation} \label{permeability}
    K(\mathbf{x}) = \mathrm{exp}(G(\mathbf{x})), \;\; G(\cdot) \sim \mathcal{N}(\bar{\mu}, k(\cdot,\cdot)) \; . 
\end{equation}
We take the covariance function as
\begin{equation} \label{eq:cov_func}
    k(\mathbf{x},\mathbf{x}') = \textrm{exp}(-\vert \vert \mathbf{x}-\mathbf{x}'\vert \vert _2/l)
\end{equation}
in our experiments, as in~\cite{Zhu_2018}. 

For dimensionality reduction and to create a parametric representation of the data, the intrinsic dimensionality $s$ of the data is specified by leveraging the Karhunen-Loeve Expansion (KLE), retaining only the first $s$ terms in
\begin{equation} \label{eq:KLE}
    G(\mathbf{x}) = \bar{\mu} + \sum_{i=1}^s \sqrt{\lambda_i} \theta_i \phi_i(\mathbf{x}) \; ,
\end{equation}
where $\lambda_i$ and $\phi_i(\mathbf{x})$ are eigenvalues and eigenfunctions of the covariance function (Eq. \ref{eq:cov_func}) sorted by decreasing $\lambda_i$, and each $\theta_i$ are sampled according to some distribution $p(\theta)$, denoted the \emph{generative parameter distribution}, which we set to be $p(\theta) = \mathcal{N}(\mathbf{0}, \mathbf{I})$. These generative parameters $\theta$ entirely determine the solution to the governing equations. In other words, the solution is deterministic function of the generative parameters. 

\subsection{Traditional Finite Difference Solution and Datasets}\label{sec:traditional_solve}
After sampling the permeability field, Eq~\ref{eq:darcy} is solved for the pressure fields. Note that once the pressure field is obtained, the velocity fields can be easily approximated using finite difference approximations using the relationship $\mathbf{u}(\mathbf{x}) = -K(\mathbf{x})\nabla p(\mathbf{x})$. We discretize the spatial domain $\mathcal{X} = [0,1]^2$ on an $n\times n$ grid to form a computational grid of $n^2$ nodes. This leads to discrete steps of $\Delta x_1 = \Delta x_2 = 1/(n-1)$ in the spatial coordinate system. Each node in the computational domain is labeled with two indices $i,j$ corresponding to the spatial location of the node such that $\mathbf{x}_{i,j} = [(i-1)/(n-1), (j-1)/(n-1)]^T$. A linear system $\mathbf{A}\mathbf{p} = \mathbf{f}$ is then formed and solved for pressure on the computational domain. To create this linear system, we define the solution $p_{i,j}$ at each point in the discretized spatial domain. The Darcy flow equations for pressure only are given by
\begin{align}
    -\nabla \cdot [K(\mathbf{x})\nabla p(\mathbf{x})] & = f_s(\mathbf{x}) \label{eq:darcy_solve_1} \\ 
    \nabla p(\mathbf{x}) \cdot \hat{\mathbf{n}}(\mathbf{x}) & = 0 \label{eq:darcy_solve_2} \\ 
    \int_{\mathcal{X}} p(\mathbf{x}) d\mathbf{x} & = 0 \; . \label{eq:darcy_solve_3}
\end{align}
Equation~\ref{eq:darcy_solve_1} can be expanded using chain rule to be
\begin{equation}
    -K(\mathbf{x})\frac{\partial^2 p(\mathbf{x})}{\partial x_1^2} - \frac{\partial K(\mathbf{x})}{\partial x_1}\frac{\partial p(\mathbf{x})}{\partial x_1} - K(\mathbf{x})\frac{\partial^2 p(\mathbf{x})}{\partial x_2^2} - \frac{\partial K(\mathbf{x})}{\partial x_2}\frac{\partial p(\mathbf{x})}{\partial x_2} = f_s(\mathbf{x})
\end{equation}
On our discretized domain, we use second order central finite differences on interior points to compute first and second order partial derivatives. For example,
\[
\left. \frac{\partial p(\mathbf{x})}{\partial x_1} \right |_{\mathbf{x}_{i,j}} \approx \frac{p_{i+1,j} - p_{i-1,j}}{2\Delta x_1}
\ \ 
\textrm{and}
\ \ 
\left. \frac{\partial^2 p(\mathbf{x})}{\partial x_1^2} \right |_{\mathbf{x}_{i,j}} \approx \frac{p_{i-1,j}-2p_{i,j}+p_{i+1,j}}{\Delta x_1^2} \; .
\]
The pressure gradient across the boundaries is zero according to Eq.~\ref{eq:darcy_solve_2}. Thus on the left ($x_1=0$) boundary for example,
\[
\left. \frac{\partial p(\mathbf{x})}{\partial x_1} \right |_{\mathbf{x}_{i,j}} \approx \frac{p_{i+1,j} - p_{i-1,j}}{2\Delta x_1} = 0\;,
\]
and thus $
p_{i+1,j} = p_{i-1,j}.$
The second order partial derivative on the left ($i=1$) boundary is therefore given by
\[
\left. \frac{\partial^2 p(\mathbf{x})}{\partial x_1^2} \right |_{\mathbf{x}_{1,j}} \approx \frac{2p_{2,j}-2p_{1,j}}{\Delta x_1^2} \; .
\]
Additionally, the integral constraint in Eq.~\ref{eq:darcy_solve_3} can be enforced by adding another row to the matrix $\mathbf{A}$, creating an over-determined system of equations. We thus solve the system $\mathbf{A}\mathbf{p} = \mathbf{f}$ where $\mathbf{A} \in \mathbb{R}^{(n^2+1)\times n^2}$, $\mathbf{p}\in \mathbb{R}^{n^2}$, and $\mathbf{f} \in \mathbb{R}^{n^2+1}$. The vectors $\mathbf{p}$ and $\mathbf{f}$ are given by Eq.~\ref{eq:p_and_f}.
\begin{equation}\label{eq:p_and_f}
    \mathbf{p} = 
    \begin{bmatrix}
        p(\mathbf{x}_{1,1}) \\
        p(\mathbf{x}_{2,1}) \\
        \vdots \\
        p(\mathbf{x}_{n,1}) \\
        p(\mathbf{x}_{1,2}) \\
        \vdots \\
        p(\mathbf{x}_{n,n})
    \end{bmatrix}
    , \; \mathbf{f}_s = 
    \begin{bmatrix}
        f_s(\mathbf{x}_{1,1}) \\
        f_s(\mathbf{x}_{2,1}) \\
        \vdots \\
        f_s(\mathbf{x}_{n,1}) \\
        f_s(\mathbf{x}_{1,2}) \\
        \vdots \\
        f_s(\mathbf{x}_{n,n}) \\
        0
    \end{bmatrix}.
\end{equation}
The matrix $\mathbf{A}$ is constructed using the finite difference formulas previously described. The exact form and a description of the discretized equations which the linear system solves are included in Sec.~\ref{app:linear_system} of the included supplementary materials.

The velocity components $\mathbf{u}(\mathbf{x})$ are computed using second order finite difference approximations on $\mathbf{p}$ to compute $\mathbf{u}(\mathbf{x}) = -K(\mathbf{x})\nabla p(\mathbf{x})$. Due to this, we train the generative model using only the permeability $K(\mathbf{x})$ and pressure field $p(\mathbf{x})$ data such that a single data sample consists of the set $\mathbf{y} = \{K(\mathbf{x}), p(\mathbf{x})\}$. 

We employ two training and test datasets throughout the work. The first training dataset contains 10,000 samples generated by solving the Darcy flow equations with an intrinsic dimensionality of $s=16$ on a discretized grid of $n=64$. The generative parameters corresponding to each sample are randomly generated according to $p(\theta) = \mathcal{N}(\mathbf{0}, \mathbf{I})$ and saved to be used in conditional experiments. An additional 1,000 samples are generated and withheld from training as a test set used only in conditional experiments. The second training and test sets are generated with an intrinsic dimensionality of $s=256$, keeping all other parameters identical to the $s=16$ dataset.

\subsection{Residual Computation}\label{sec:residual}
Implementing the score-based generative model requires a method of evaluating the degree to which generated samples satisfy the Darcy flow equations. One way of doing this is to evaluate the PDE residual (Eq.~\ref{eq:darcy_residual}) on generated samples. Note that the integral condition of Eq.~\ref{eq:darcy_solve_3} can be easily satisfied for any generated sample by regularizing the intermediate output $\tilde{\mathbf{p}}$ according to
\[
\mathbf{p} = \tilde{\mathbf{p}} - \int_{\mathcal{X}} \tilde{\mathbf{p}} d\mathbf{x} \; .
\]
We thus ignore this as part of the residual computation and satisfy the integral condition by construction. The residual is therefore considered only to be
\begin{equation}\label{eq:darcy_residual}
    \mathbf{r} = K(\mathbf{x})\frac{\partial^2 p(\mathbf{x})}{\partial x_1^2} + \frac{\partial K(\mathbf{x})}{\partial x_1}\frac{\partial p(\mathbf{x})}{\partial x_1} + K(\mathbf{x})\frac{\partial^2 p(\mathbf{x})}{\partial x_2^2} + \frac{\partial K(\mathbf{x})}{\partial x_2}\frac{\partial p(\mathbf{x})}{\partial x_2} + f(\mathbf{x}) \; ,
\end{equation}
following Eq.~\ref{eq:darcy_solve_1}. This residual is a function of the spatial coordinate $\mathbf{x}$ and can be approximated at each point in the computational domain given $K(\mathbf{x})$ and $p(\mathbf{x})$. However, when applying physical consistency steps according to Eq.~\ref{eq:physical_consistency}, we apply the residual gradient only to pressure fields, avoiding direct updates of the permeability field.

When solving the Darcy flow equations as outlined above, the gradients $\partial p/\partial x_1 = 0$ and $\partial p/\partial x_2 = 0$ are approximately satisfied on the domain boundaries by construction of the linear system. However, in computing the residual, we must compute gradients on these boundaries. We therefore use second order forward and backward difference approximations to the first and second order derivatives on the boundaries to compute the residual. This facilitates approximating the residual on both data generated by solving the linear system $\mathbf{A}\mathbf{p}=\mathbf{f}$ and samples obtained from the generative model by solving the reverse SDE.

\subsection{Physically-Consistent Unconditional Generative Model}\label{sec:darcy_uncond}
Constraining generated samples to follow physical equations as described in Sec.~\ref{sec:physical_consistency} is quite flexible and inherently contains hyperparameters which can be set and altered after training. In particular, the number of discrete time steps $\tau$ used in the solver, the number of residual minimization steps $N$ prior to $T=0$, and the number of residual minimization steps $M$ after solving the reverse SDE or reverse PF ODE all play an important role in enforcing the physical laws which samples are to follow. We demonstrate our method of enforcing physical consistency by training an unconditional generative model on the Darcy flow training dataset with dimensionality $s=16$.

\subsubsection{Training}


Our unconditional model architecture is illustrated as the left side of Fig~\ref{fig:controlnet_arch}. Each of the components labeled with parameters $\phi$ constitute the unconditional generative model. Additional architectural, training, and hyperparameter details are provided in Sec.~\ref{app:training_uncond} of the included supplementary materials.

\subsubsection{Physically-Consistent Sampling}
After training, the sampling process is quite flexible as discussed in~\ref{sec:physical_consistency_sampling}. Sampling amounts to solving Eq.~\ref{eq:sde_reverse_approx} or~\ref{eq:probability_flow_ode} backwards in time augmented with physical consistency steps, but the particular method of solving has a large impact on the quality of the final samples at $t=0$. We investigate the effect that each equation and various combinations of hyperparameters have on the average physical residual for generated samples. 

Higher-order solvers can also be used to provide gains in sampling efficiency~\cite{karras2022elucidating}, but is outside the scope of this work. Instead, we aim to minimize sample residuals without an initial emphasis on sampling efficiency. The use of physical consistency steps defined in Sec.~\ref{sec:physical_consistency_sampling} along with the number of time steps used to solve either the reverse SDE or PF ODE constitute our investigation into reducing the physical residual of generated samples. 

In terms of computational cost, the physical residuals are computed using finite differences, which are relatively inexpensive compared to the cost of the forward pass of the diffusion model. On an Nvidia A6000 GPU, the average time taken for solving one PF ODE step with a batch size of one is approximately \(3.1 \times 10^{-3}\) seconds, while the average time taken for evaluating the physical residual is about \(1.5 \times 10^{-5}\) seconds. The SDE formulation used in this work takes $\mathcal{O}(1000)$ steps in the reverse sampling process. However, there are other SDE formulations~\cite{esser2024scaling} with straightened flow trajectories that can significantly reduce the number of sampling steps required. These alternative formulations can achieve comparable performance with fewer iterations, thereby enhancing computational efficiency without sacrificing accuracy.


When implementing physical consistency steps to reduce the PDE operator residual, we exclusively apply the steps to the pressure component of the samples. This deliberate choice ensures that no adjustments are made to the permeability field. Since the pressure field is a deterministic function of the permeability field, we refrain from directly modifying the permeability field generated by the model. Instead, we allow the model to fully generate the permeability field without interference.

\begin{figure}[h!]
    \centering
    \includegraphics[width=.75\linewidth]{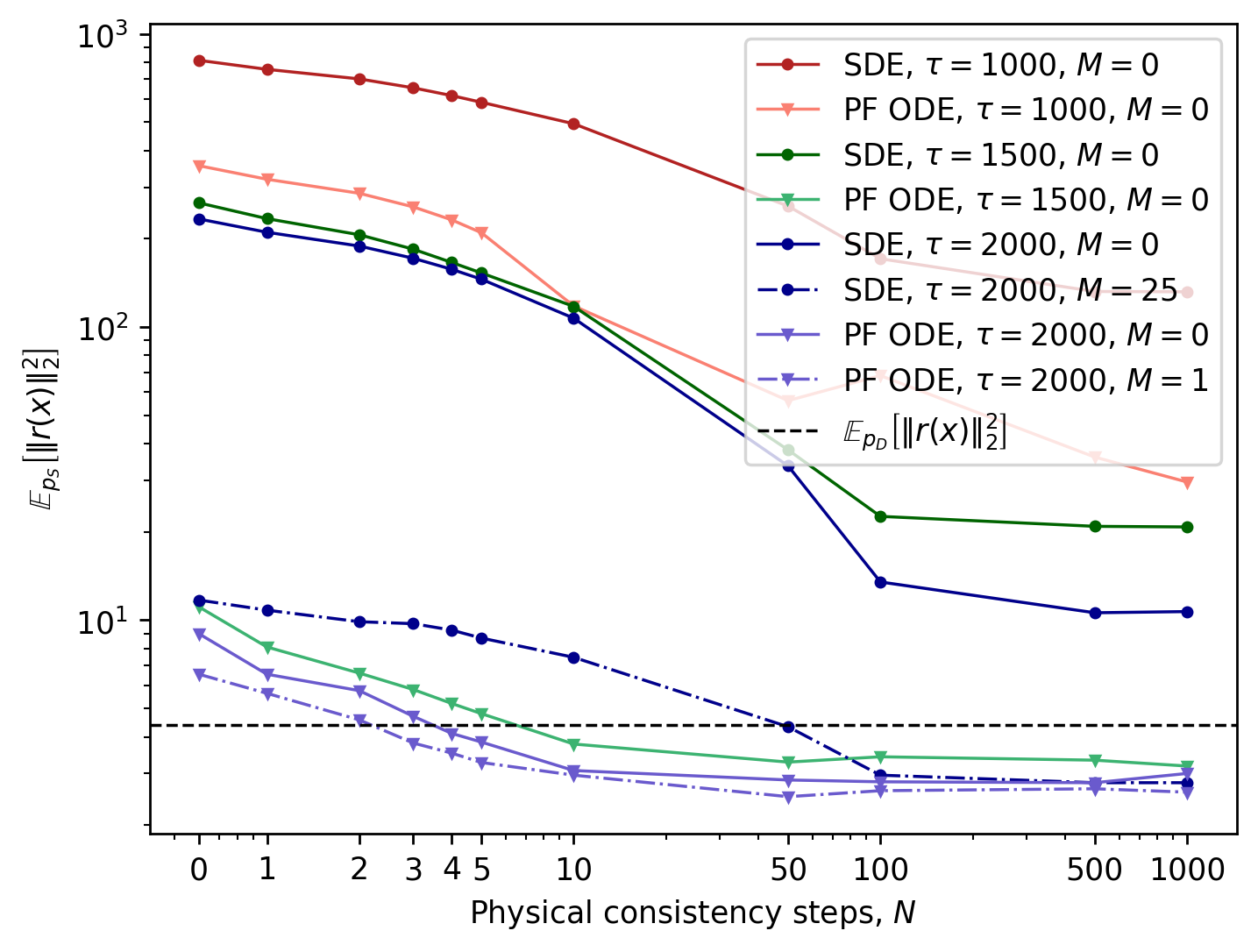}
    \caption{Darcy flow equations are enforced through physical consistency sampling. Hyperparameters $\tau$, $N$, and $M$ as well as the sampling equation have a large impact on physical residuals.}
    \label{fig:consistency_s16}
\end{figure}

Sampling is performed by varying $\tau$, $N$, and $M$ and solving either the PF ODE or the reverse SDE to generate a synthetic dataset of 1,000 samples. For each dataset, the average physical residual is computed as outlined in Sec.~\ref{sec:residual}. The primary results of this experiment are illustrated in Fig.~\ref{fig:consistency_s16}, where $p_D(x)$ denotes the training data distribution and $p_S(x)$ denotes the distribution of 1,000 samples generated according to the indicated method.

Initially, we observe that the choice of the number of discrete timesteps $\tau$ is crucial when employing the Euler-Maruyama method for solving the reverse SDE or the forward Euler method for solving the PF ODE. Insufficiently small values of $\tau$ result in high errors in approximating the dynamics described by Eqs.~\ref{eq:VP_reverse_SDE} and~\ref{eq:VP_reverse_PFODE}, leading to suboptimal sampling quality and increased physical residuals. In both scenarios, solving either the reverse SDE or PF ODE with a value of  $\tau$ which is too small renders physical consistency enforcement incapable of reducing residuals in generated samples to match the quality of the training dataset.

However, with a sufficiently large value of $\tau$, physical consistency can effectively reduce residuals on the generated datasets to be equal to or even lower than those in the training dataset. In the context of solving the reverse SDE, the stochastic nature of the process necessitates large numbers of physical consistency steps to reduce residuals towards the training dataset levels. However, when solving the PF ODE, relatively fewer physical consistency steps are required. The PF ODE generally yields significantly smaller residuals on generated samples compared to the reverse SDE, regardless of the presence of physical consistency steps. The additional noise provided by solving the reverse SDE may be beneficial in some cases in which the importance of diversity may outweigh that of fidelity. However, in many physical problems where fidelity is paramount, the PF ODE emerges as the superior choice.

The minimum number of PDE operator evaluations needed to achieve a generated dataset residual less than or equal to that of the training set was attained by solving the PF ODE with $\tau = 2000$, $M=1$, and $N=3$, resulting in a total of 4 evaluations.

\begin{figure}[h!]
    \centering
    \begin{subfigure}[b]{0.42\textwidth}
        \includegraphics[width=\linewidth]{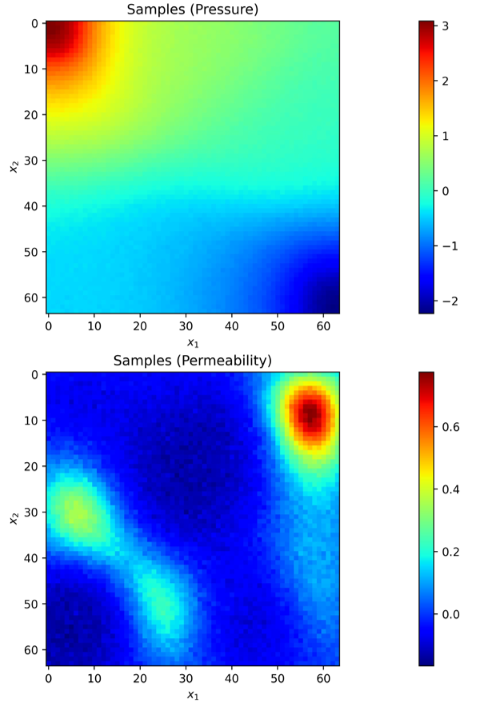}
        \caption{$N=0$}
    \end{subfigure}
    \begin{subfigure}[b]{0.42\textwidth}
        \includegraphics[width=\linewidth]{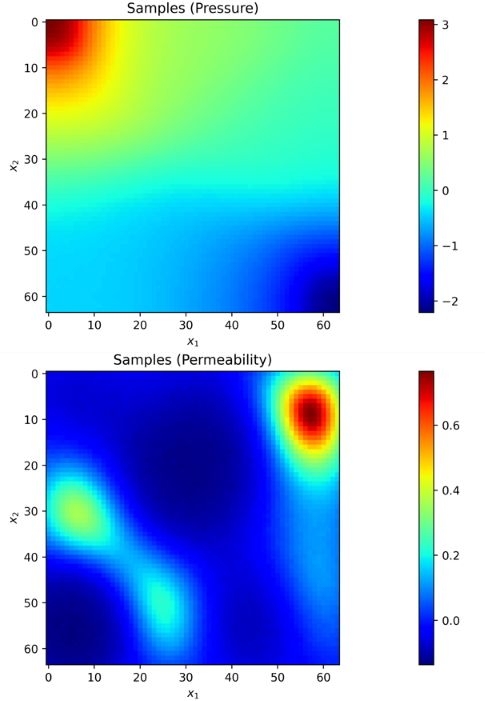}
        \caption{$N=50$}
    \end{subfigure}
    \caption{Physical consistency steps applied only to pressure fields provide a denoising effect on permeability fields while reducing the physical residual. Samples are obtained by solving the reverse SDE using EM solver with $\tau=2000$ time steps and $M=0$ additional physical consistency steps.}
    \label{fig:darcy_sde_N_comp}
\end{figure}


We observe that the number of physical consistency steps, denoted as $N$, preceding the time $t=0$ serves a dual purpose beyond reducing the physical residual. As mentioned, we exclusively apply physical consistency steps to pressure fields.  Given that the score function approximation jointly predicts the score functions for both pressure and permeability fields, a correlation emerges in the denoising process for each field. Consequently, directly modifying the pressure field during the reverse process indirectly influences the permeability field through the score function prediction. This, in turn, produces a denoising effect on the permeability fields while concurrently reducing the physical residual. Figure~\ref{fig:darcy_sde_N_comp} illustrates the denoising effect that physical consistency steps have on pressure and permeability fields. Applying physical consistency steps provides two main benefits: encouraging generated samples to follow the physical equations of interest as well providing a denoising effect during sampling. Such benefits are not exclusive to unconditional models and can also be applied to conditional sample generation for forward and inverse problems. 

%% file: 05_conditional.tex
\section{Conditional Generation}\label{sec:conditional}

Conditional generative models hold immense potential in physics-based applications, offering versatile utility in various tasks such as field prediction, uncertainty quantification, and field reconstruction. Their broad range of applications makes it advantageous to train a single unconditional generative model that can be leveraged by numerous conditional models.

For instance, in the context of field prediction, conditional generative models can be trained to generate specific fields based on various input conditions such as generative parameters, boundary conditions, partial field measurements, or other relevant factors, effectively solving a forward prediction problem. This allows for the flexible generation of fields tailored to specific scenarios. 
Field reconstruction involves the task of reconstructing a complete field based on partial or incomplete information, and constitutes a reverse problem. Conditional generative models can be trained to reconstruct fields by leveraging available data and conditioning information, offering a powerful tool in scenarios where complete field data may be limited or noisy. In both cases of solving a forward or inverse problem, physical consistency sampling ensures that all predictions will adhere to the underlying physical behavior governed by the PDE. 

The advantage of training a single unconditional generative model lies in its ability to capture the underlying distribution of the data. This model can then be flexibly adapted for various conditional tasks, eliminating the need to train separate models for each specific condition. This approach not only streamlines the training process but also facilitates efficient model reuse across a diverse set of applications.

We train two separate conditional augmentations on the Darcy flow dataset with $s=16$, each of which are connected to the unconditional model architecture as illustrated in Fig.~\ref{fig:controlnet_arch}. The conditional portion of the model (blue) can be removed from the overall architecture to leave behind an unconditional model. This renders it easy to add or remove different conditional augmentations to the same pretrained unconditional model. We copy the same unconditional model trained in Sec.~\ref{sec:unconditional}, and train two separate augmentations as described in Sec.~\ref{sec:controlnet}: one for pressure and permeability field prediction from the underlying generative parameters of the permeability field and one for pressure / permeability field reconstruction from partial pressure measurements.

\subsection{Conditional Field Generation from Generative Parameters}\label{sec:surrogate_model}
\begin{figure}[h!]
    \centering
    \includegraphics[width=1.0\linewidth]{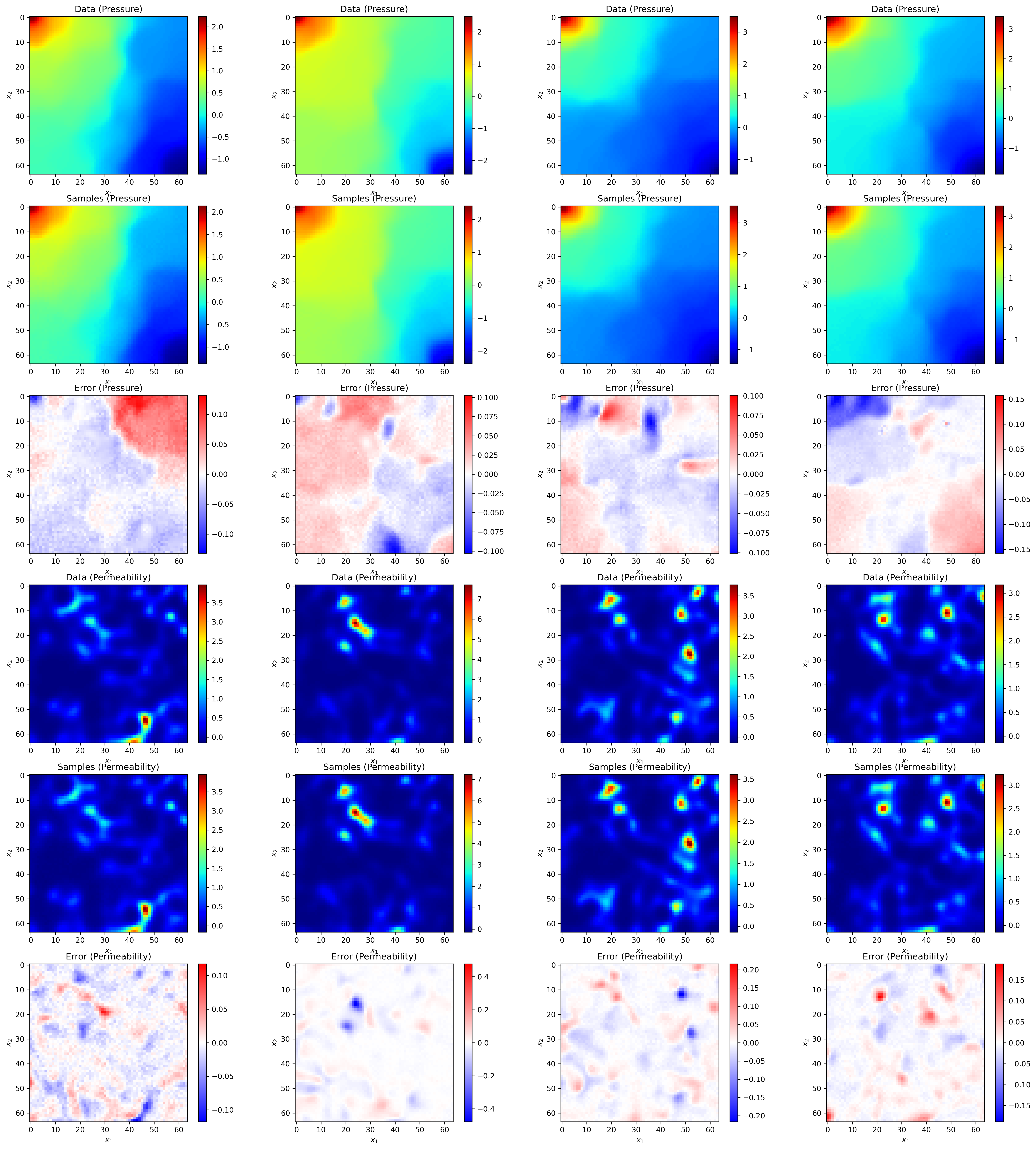}
    \caption{Samples from a conditional model trained on Darcy flow data ($s=256$) where the conditional augmentation takes permeability field generative parameters as input.}
    \label{fig:darcy_controlnet_params_256}
\end{figure}
We illustrate the first application for conditional generative models as that of surrogate modeling, a forward problem. A conditional augmentation is trained with a baseline unconditional model on the Darcy flow dataset with an underlying dimensionality of $s=256$. The conditioning input $\theta \in \mathbb{R}^s$ are the parameters $\theta$ determining the permeability field in Eq.~\ref{eq:KLE}. The conditional generative model will output samples from $p(\mathbf{y}|\theta)$, given the particular parameterization. As data samples $x$ are determined by an injective mapping $G(\theta): \mathbb{R}^s \rightarrow \mathbb{R}^{n^2}$, the true conditional p.d.f. is zero everywhere except $\mathbf{y} = G(\theta)$. 

The training procedure is outlined in Sec.~\ref{sec:controlnet}, and additional details are included in Sec.~\ref{app:training_cond} of the included supplementary materials. In this case, the three control encoding blocks in Fig~\ref{fig:controlnet_arch} are identical in structure to the encoding blocks of the unconditional U-Net~\cite{10.1007/978-3-319-24574-4_28} with the exception of a zero-convolution at the output of each block. The input to the first control encoding block must therefore be the same shape as the noisy inputs $\mathbf{y}(t)$. To achieve this, the condition encoder consists of linear layers which encode the conditioning from $\mathbb{R}^s$ to $\mathbb{R}^{n^2}$ and reshape it to be directly summed with $\mathbf{y}(t)$. 

After training, sampling is performed using conditions $\theta$ from a separate test set which were not present in the training data. We sampling with physical consistency enforcement by solving the PF ODE with $\tau = 2000$, $N=50$, and $M=10$. Both qualitative and quantitative reconstructions are provided in Fig~\ref{fig:darcy_controlnet_params_256}, illustrating accurate predictions for both pressure and permeability fields from the underlying parameterization of the permeability field. Predictions on the test set result in an average $L_2$-norm of $7.0\times 10^{-4}$ on pressure field predictions and $6\times 10^{-4}$ on permeability field predictions. We note that prediction is performed by drawing a single sample from $p(\mathbf{y}|\theta)$ for each value of $\theta$ in the test set. 


\subsection{Conditional generation for temporal PDE}

To ensure broader applicability of the method, the 1D Burgers equation is considered as a representative example of a Hyperbolic PDE:
\begin{equation}
\label{eq:burgers}
\frac{\partial u}{\partial t} + u \frac{\partial u}{\partial x} = \nu \frac{\partial^2 u}{\partial x^2},
\end{equation}
where $u$ is the velocity field, $\nu$ is the viscosity coefficient, and $x$ is the spatial coordinate. Spatially periodic boundary conditions are used and the initial condition is assumed to be a superposition of sinusoidal waves, using the formulation provided in PDEBench~\cite{takamoto2022pdebench}:
\begin{equation}
\label{eq:burgers_ic}
u_0(x) = \sum_{k_i=k_1,...,k_N} A_i \sin(k_i + \phi_i),
\end{equation}

\noindent where $A_i$ is the amplitude, selected from a uniform distribution $[0, 1]$, $k_i$ is the wave number, with $k_{max}=8$, $N=2$, and $\phi_i$ is the phase, selected from a uniform distribution $[0, 2\pi]$.

The dataset is generated based on the script provided in PDEBench \cite{takamoto2022pdebench} with $\nu=0.01$. The grid size is 64, and time step 0.01 for 64 steps. The dataset can be interpreted as time slabs rather than time steps, thus treated as 2D data, with spatial and temporal dimensions. We select the conditioning input, $\theta$, to be the initial condition, $u_0(x)$, and the target output, $\mathbf{y}$, to be the velocity field at each time step, $u(x, t)$. This formulation is analogous to solving the PDE with the initial and boundary conditions.

The discretized residual equation of the 1D Burgers equation can be written as:
\begin{equation}
\label{eq:burgers_residual}
\mathbf{r}_{i,j} = \frac{u_{i,j+1} - u_{i,j-1}}{2 \Delta t} + u_{i,j} \frac{u_{i+1,j} - u_{i-1,j}}{2 \Delta x} - \nu  \frac{u_{i+1,j} - 2u_{i,j} + u_{i-1,j}}{\Delta x^2}
\end{equation}
Similar to the residual computation for the Darcy flow problem, the time derivatives at the boundaries are calculated using second-order forward and backward finite differences. For temporal problems, the time step used in the simulation may be different from the time step in the dataset, thus, the computed residual may not be precisely zero. In this example, we pick the residual step size, $\epsilon$, to be $5\times 10^{-3}$ and vary the number of physical consistency steps from 0 to 500 with changing $M$ from 0 to 5.

\begin{figure}[htb!]
    \centering
    \begin{subfigure}[b]{\textwidth}
        \includegraphics[width=\textwidth]{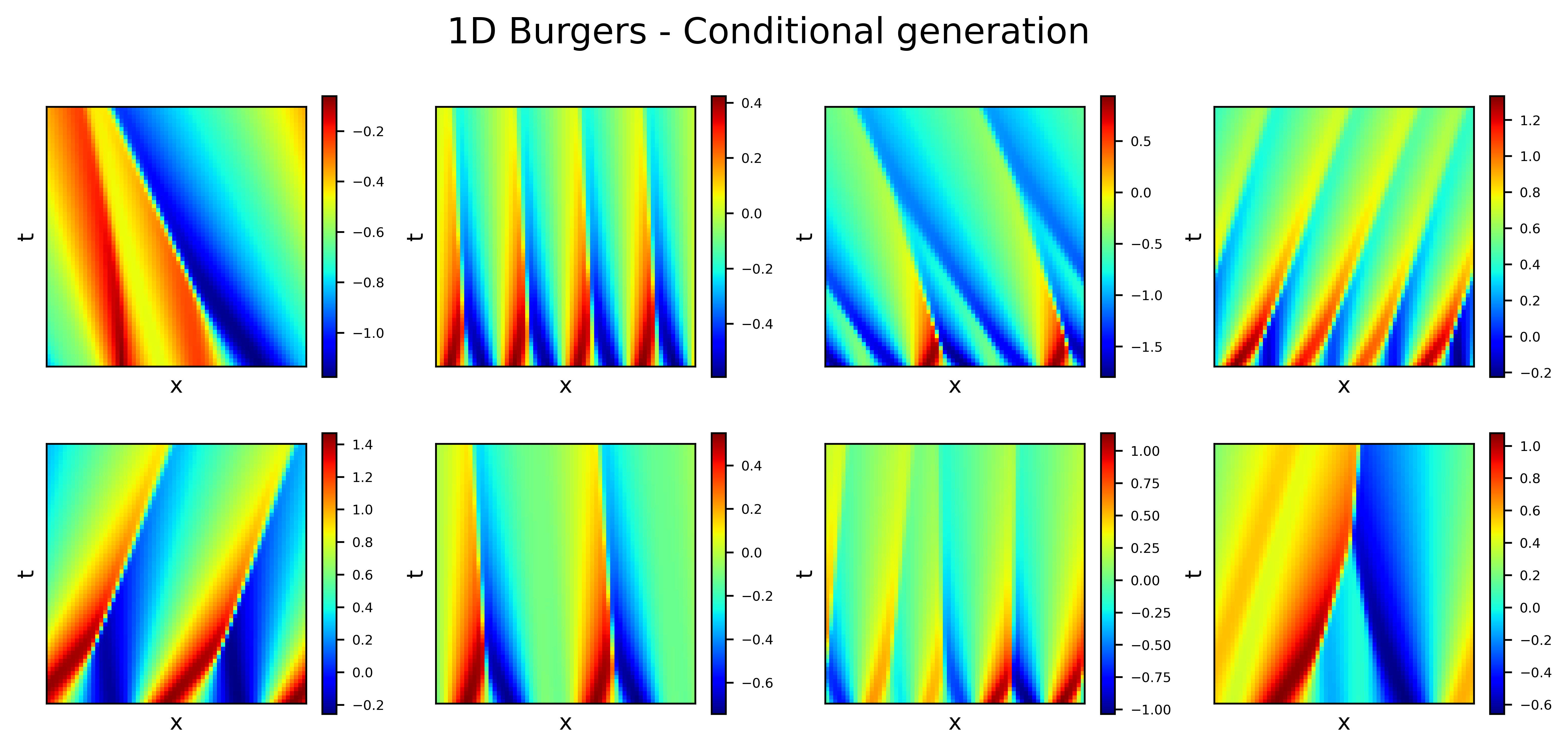}
    \end{subfigure}
    
    \begin{subfigure}[b]{\textwidth}
        \includegraphics[width=\textwidth]{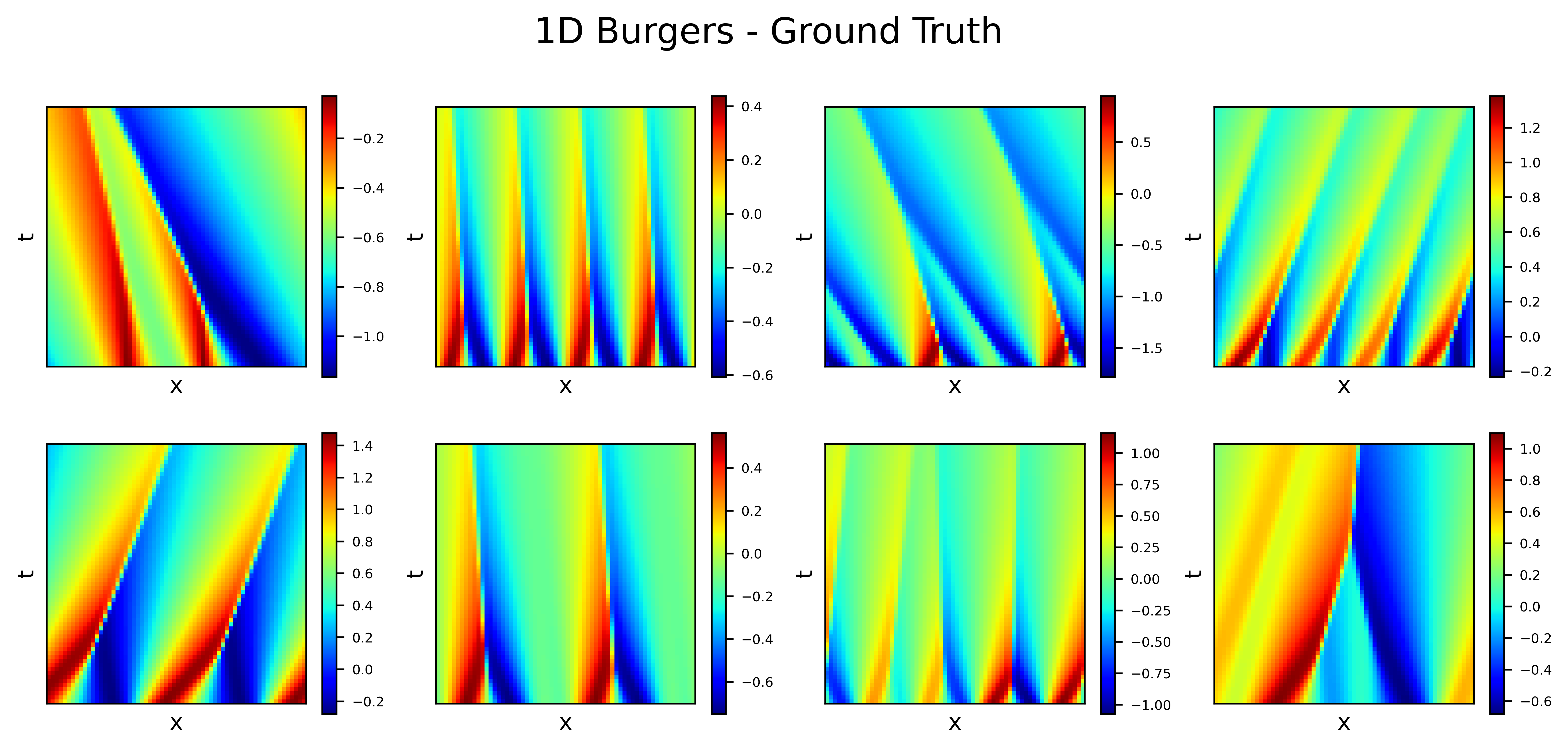}
    \end{subfigure}
    
    \begin{subfigure}[b]{\textwidth}
    \end{subfigure}
    \caption{Comparison of the conditional generation results for the 1D Burgers equation. The top row shows the generated velocity field with 18 physical consistency steps, and the bottom row shows the ground truth velocity field. The eight initial conditions are randomly selected from the test dataset.}
    \label{fig:1D_Burgers_Conditional_generation}
\end{figure}

\begin{figure}[htb!]
    \centering
    \includegraphics[width=\textwidth]{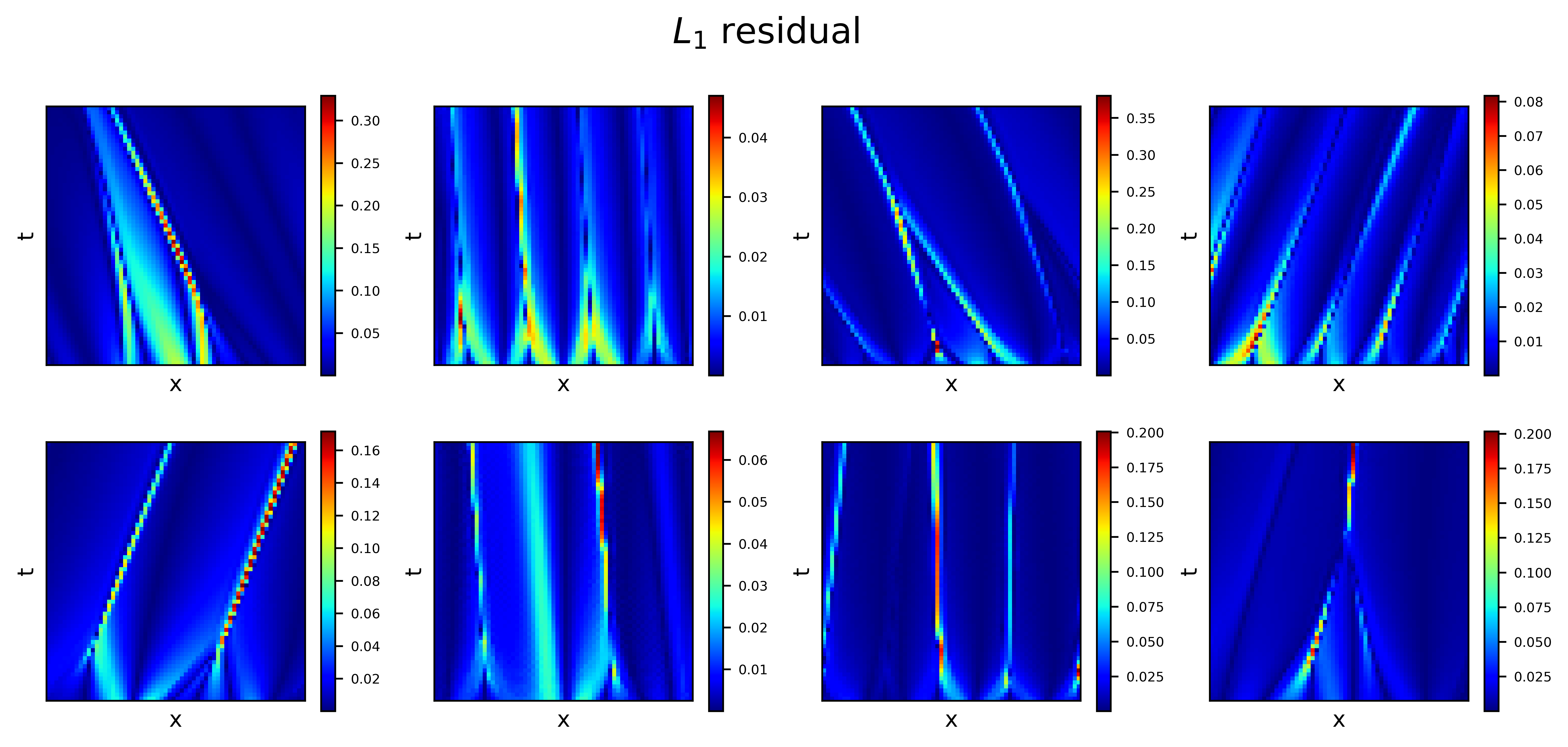}
    \caption{$L_1$ norm of the residual of the eight selected simulations for the 1D Burgers equation with 18 physical consistency steps.}
    \label{fig:1D_Burgers_L1_residual}
\end{figure}

The dataset consists of 10,000 simulations, with a split ratio of 0.7:0.2:0.1 for training, validation, and testing. The diffusion model employs the classifier-free guidance~\cite{ho2022classifier} for handling the conditioning input. For details of the implementation and the training procedure, please refer to the Appendix~\ref{app:edm}.

Fig~\ref{fig:1D_Burgers_Conditional_generation} shows the conditional generation results for the 1D Burgers equation. The eight initial conditions are randomly selected from the test dataset. The results demonstrate that the model can accurately generate the velocity field, indicating that the model effectively captures the underlying advection and diffusion processes. The $L_1$ norm of the residual, computed from the difference between the generated velocity field and the ground truth, is shown in Fig~\ref{fig:1D_Burgers_L1_residual}. The sampled solution can capture the sharp interface of the shock wave, and the residual is mainly due to the position of the shock wave. Since the velocity profiles at each time step are generated simultaneously, the errors do not propagate over the temporal dimension over the given time steps. This property could help stabilize the unrolled prediction for future time steps by using the last time step as the conditioning input.

\begin{table}[h!]
\centering
\caption{Comparison of RMSE and physical residual for the 1D Burgers equation using CoCoGen with different numbers of physical consistency steps.}
\label{tab:CoCoGen_RMSE}
\begin{tabular}{|l|cccccc|}
\hline
Phy. Steps & \multicolumn{1}{c|}{0}         & \multicolumn{1}{c|}{100}         & \multicolumn{1}{c|}{200}         & \multicolumn{1}{c|}{300}        & \multicolumn{1}{c|}{400}        &  500        \\ \hline
RMSE       & \multicolumn{1}{c|}{3.69E-2} & \multicolumn{1}{c|}{3.46E-2} & \multicolumn{1}{c|}{3.45E-2} & \multicolumn{1}{c|}{3.46E-2} & \multicolumn{1}{c|}{3.51E-2} & 3.61E-2  \\ \hline
Phy. Res.  & \multicolumn{1}{c|}{8.84}    & \multicolumn{1}{c|}{7.52}    & \multicolumn{1}{c|}{7.25}    & \multicolumn{1}{c|}{6.95}    & \multicolumn{1}{c|}{6.60}    & 6.24     \\ \hline
\end{tabular}
\end{table}

\begin{table}[h!]
\centering
\caption{Comparison of RMSE for the 1D Burgers equation using U-Net, FNO, and PINN.}
\label{tab:Baselines_RMSE}
\begin{tabular}{|l|c|c|c|}
\hline
Model & U-Net$^{\footnotemark[1]}$ \cite{takamoto2022pdebench} & FNO$^{\footnotemark[1]}$ \cite{takamoto2022pdebench} & PINN$^{\footnotemark[1]}$ \cite{takamoto2022pdebench} \\ \hline
RMSE  & 9.7E-2  & 6.4E-3 & 5.3E-1 \\ \hline
\end{tabular}
\end{table}
\addtocounter{footnote}{1}
\footnotetext[1]{Results taken from the autoregressive baseline models trained on 1D Burgers equation with $\nu=0.01$.}

Tab~\ref{tab:CoCoGen_RMSE} presents the comparison of the root mean square error (RMSE) and physical residual for the 1D Burgers equation with different numbers of physical consistency steps. In this example, performing more physical consistency steps reduces the physical residual and slightly improves the RMSE. Tab~\ref{tab:Baselines_RMSE} shows the RMSE of three baselines models reported in PDEBench~\cite{takamoto2022pdebench}. The RMSE of the proposed model is comparable to these autoregressive baseline models. The dataset used in this example is generated from the same script but with a smaller spatial grid size and fewer time steps. The RMSE of the baseline models is computed from single-step predictions, whereas the diffusion model can be viewed as performing 63 steps of unrolled prediction simultaneously. The results indicate that the diffusion model could potentially provide a new perspective for performing forward tasks for temporal PDEs.

\subsubsection{Conditional Field Inversion and Reconstruction with Sparse Measurements}\label{sec:firms}

%
\begin{figure}[h!]
    \centering
    \includegraphics[width=0.65\linewidth]{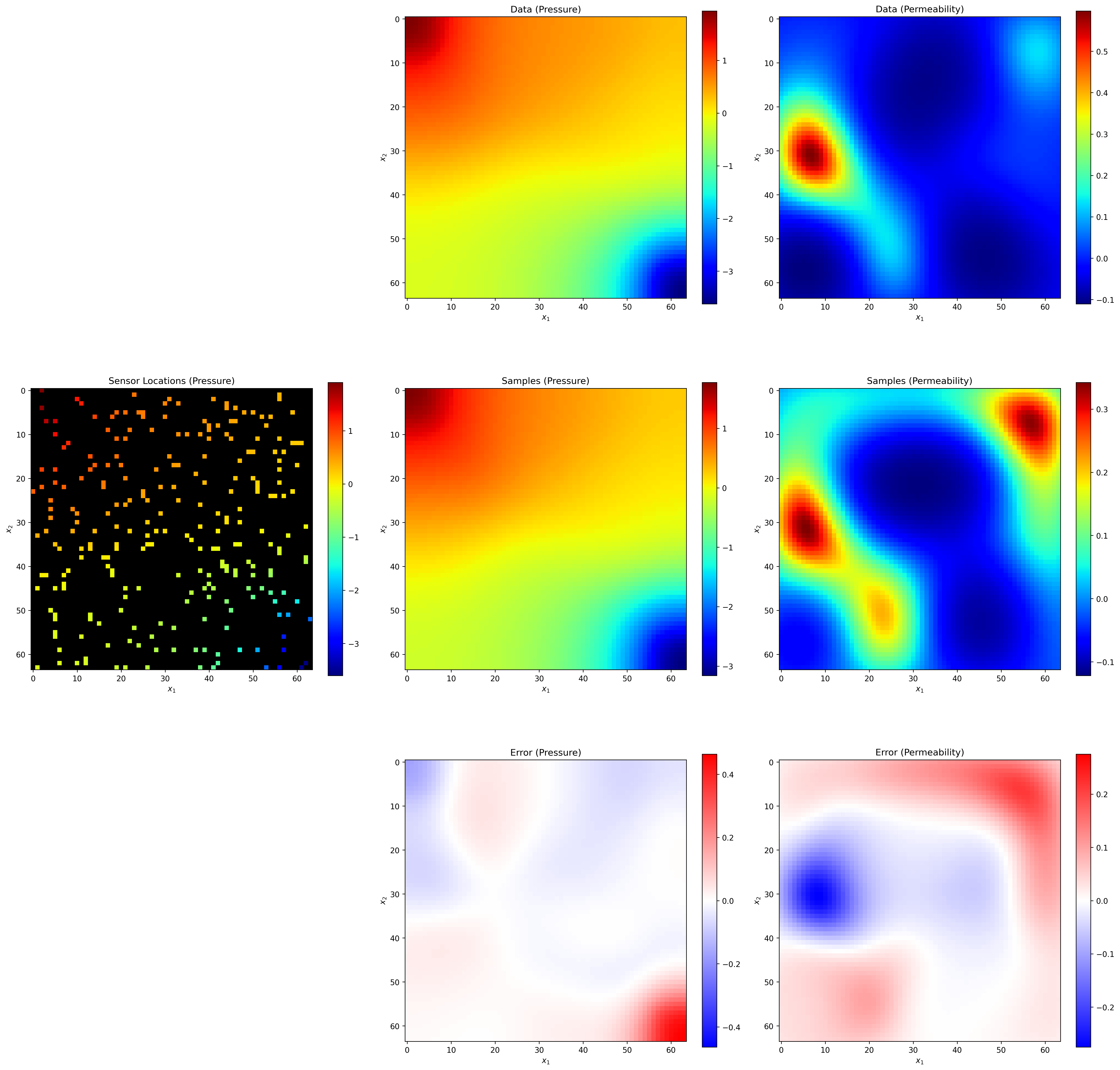}
    \caption{Sampling from $p(\mathbf{y}|\theta)$ where $\theta$ are $m=250$ partial measurements of the pressure field. Samples are always consistent with the underlying PDE, even if prediction accuracy is low.}
    \label{fig:firms_single_prediction}
\end{figure}

In addition to solving forward problems, we also illustrate the application of score-based generative models to solve inverse problems. We train a conditional augmentation to perform probabilistic field inversion and reconstruction from sparse measurements. The goal is to perform field inversion to obtain permeability field approximations and reconstruction to obtain pressure field approximations from only measurements of the pressure field. During training, we select a random number $m$ of pressure sensors, place them randomly in the physical domain, and assume accurate pressure measurements at each location. These pressure measurements are taken as the conditioning variables $\theta$ such that we sample from $p(\mathbf{y}|\theta)$ given the measurements. Further training details are provided in Sec.~\ref{app:training_cond} of the included supplementary materials.

\begin{figure}[h!]
    \centering
    \includegraphics[width=0.8\linewidth]{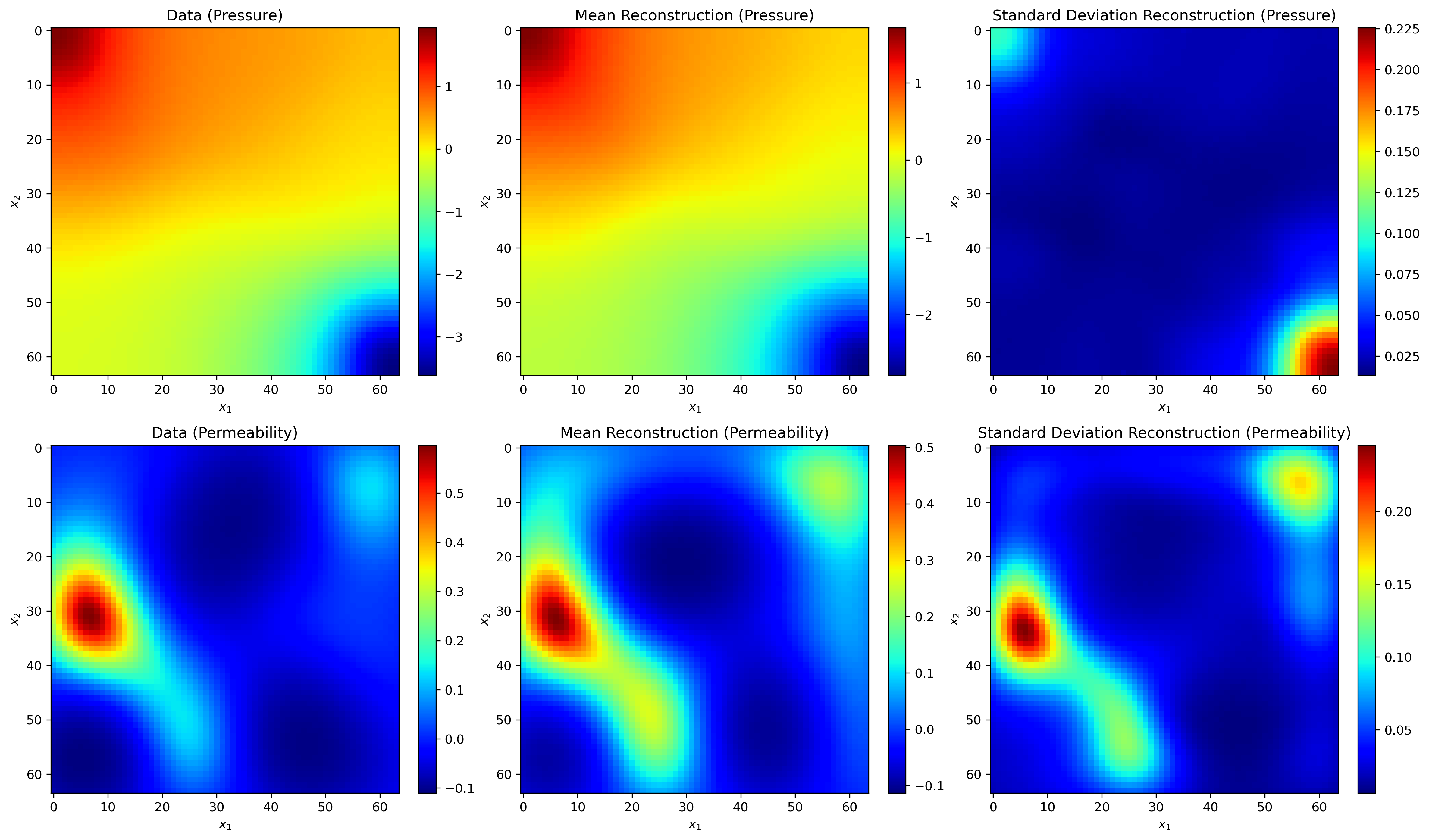}
    \caption{Many samples can be drawn from $p(\mathbf{y}|\theta)$, facilitating uncertainty quantification. The data sample (left) is compared to the expectation $\mathbb{E}_{p(\mathbf{y}|\theta)}[\mathbf{y}]$ (center) and standard deviation (right) for both pressure (top) and permeability (bottom) fields.} 
    \label{fig:firms_uncertainty}
\end{figure}

Figure~\ref{fig:firms_single_prediction} illustrates a single sample drawn from $p(\mathbf{y}|\theta)$ along with the conditioning information (pressure measurements with $m=250$ sensors) and ground truth. However, as the generative model provides the ability to sample from the conditional distribution, uncertainty quantification is possible when predicting the pressure and permeability fields. This is illustrated in Fig.~\ref{fig:firms_uncertainty}, showing the mean and standard deviation of $p(\mathbf{y}|\theta)$ compared to the true data sample. In reality, the conditional distribution is likely not a Gaussian distribution, but is very flexible due to the use of score-based generative models. We visualize only the mean and standard deviation in Fig.~\ref{fig:firms_uncertainty} for simplicity. In contrast to the surrogate modeling example in Sec.~\ref{sec:surrogate_model}, it is difficult to guarantee that $\mathbf{y}$ is a deterministic function of $\theta$. For example, consider a case in which a single pressure measurement is given. There exists no unique solution to the entire pressure and permeability fields which will exhibit the correct pressure at the single sensor location. Therefore, uncertainty quantification may be very useful in a real-world situation to determine the reliability or capacity of the current measurements to fully determine the physical fields. Further, as sampling is always performed with physical consistency enforcement (PF ODE, $\tau=2000$, $N=50$, $M=10$), all samples will approximately adhere to the governing PDE. 

\begin{figure}[h!]
    \centering
    \includegraphics[width=0.6\linewidth]{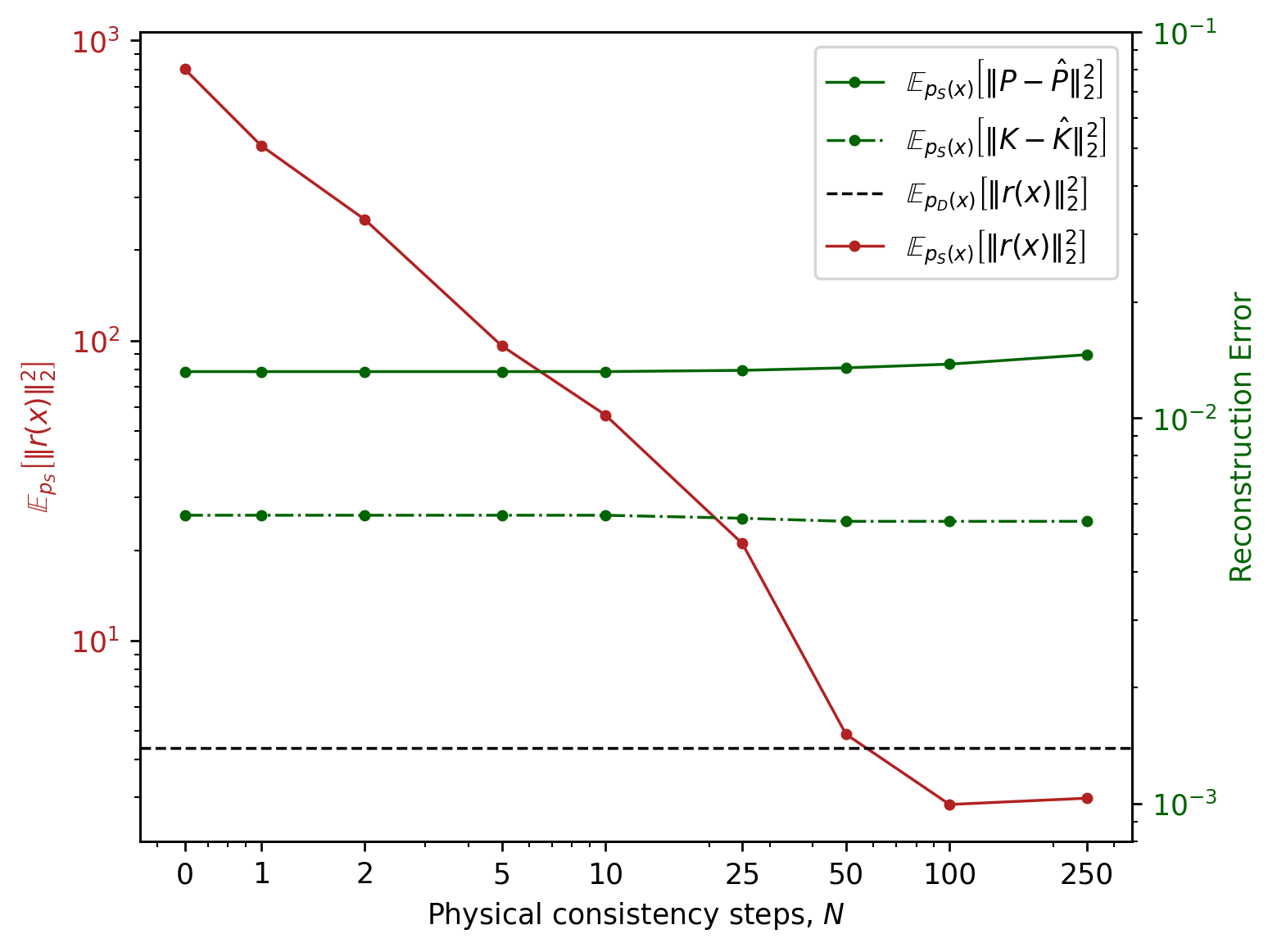}
    \caption{Generating conditional samples with physical consistency enforcement for field inversion and reconstruction with sparse measurements ($m=250$) has minimal impact on reconstruction / inversion errors while improving sample consistency with the physical PDE.}
    \label{fig:physical_consistency_recon_16}
\end{figure}

In an effort to quantify the average field reconstruction error for the case illustrated in Fig.~\ref{fig:firms_single_prediction}, we apply the same $m=250$ pressure measurements to each of the 1,000 cases in the test set, predict the pressure and permeability fields, and measure the error. The pressure reconstruction error is computed as $\mathbb{E}_{p(x|\theta)}[\| P - \hat{P} \| ^2_2]$, and the permeability reconstruction error is computed as $\mathbb{E}_{p(x|\theta)}[\| K - \hat{K} \| ^2_2]$. The pressure field approximation $\hat{P}$ and permeability field approximation $\hat{K}$ are compared to the true pressure $P$ and permeability $K$ fields. This expectation is approximated by sampling only a single time from $p(x|\theta)$, essentially sampling from the model a single time for each data sample. Although Fig.~\ref{fig:firms_uncertainty} illustrates that uncertainty may be high in $p(x|\theta)$, we use this as an illustrative example which does not require a thorough estimation of the expectation. 

Our first objective is to illustrate the effect that physical consistency sampling has on the reconstructive performance of the model. Figure~\ref{fig:physical_consistency_recon_16} shows that as the number of physical consistency steps $N$ is increased, there is nearly zero effect on the reconstructive performance until a large number of steps are used. Even then, in the worst case investigated, only a 2\% increase in pressure reconstruction error is observed while permeability reconstruction remains constant. However, notably the physical residual decreases consistently with increasing number of physical consistency steps. This shows that physics can be enforced without negatively impacting the reconstructive performance on the model, even though physics are not enforced during training. One possible explanation for this is that the measurements themselves are included as conditioning. Although PDE information is not included at training time, the measurement information is. As the sampling process is performed, iteratively passing the sample through the score estimation model, physical consistency steps will ultimately result in small changes to the sample before the sampling process is complete. The samples including these small changes are then input into the score estimation function along with the measurement information as conditioning, allowing the model to adapt itself to the small changes caused by physical consistency sampling. In effect, this facilitates the reduction of residuals while maintaining constant reconstruction error.

\begin{figure}[h!]
    \centering
    \begin{subfigure}[b]{0.45\textwidth}
        \includegraphics[width=\linewidth]{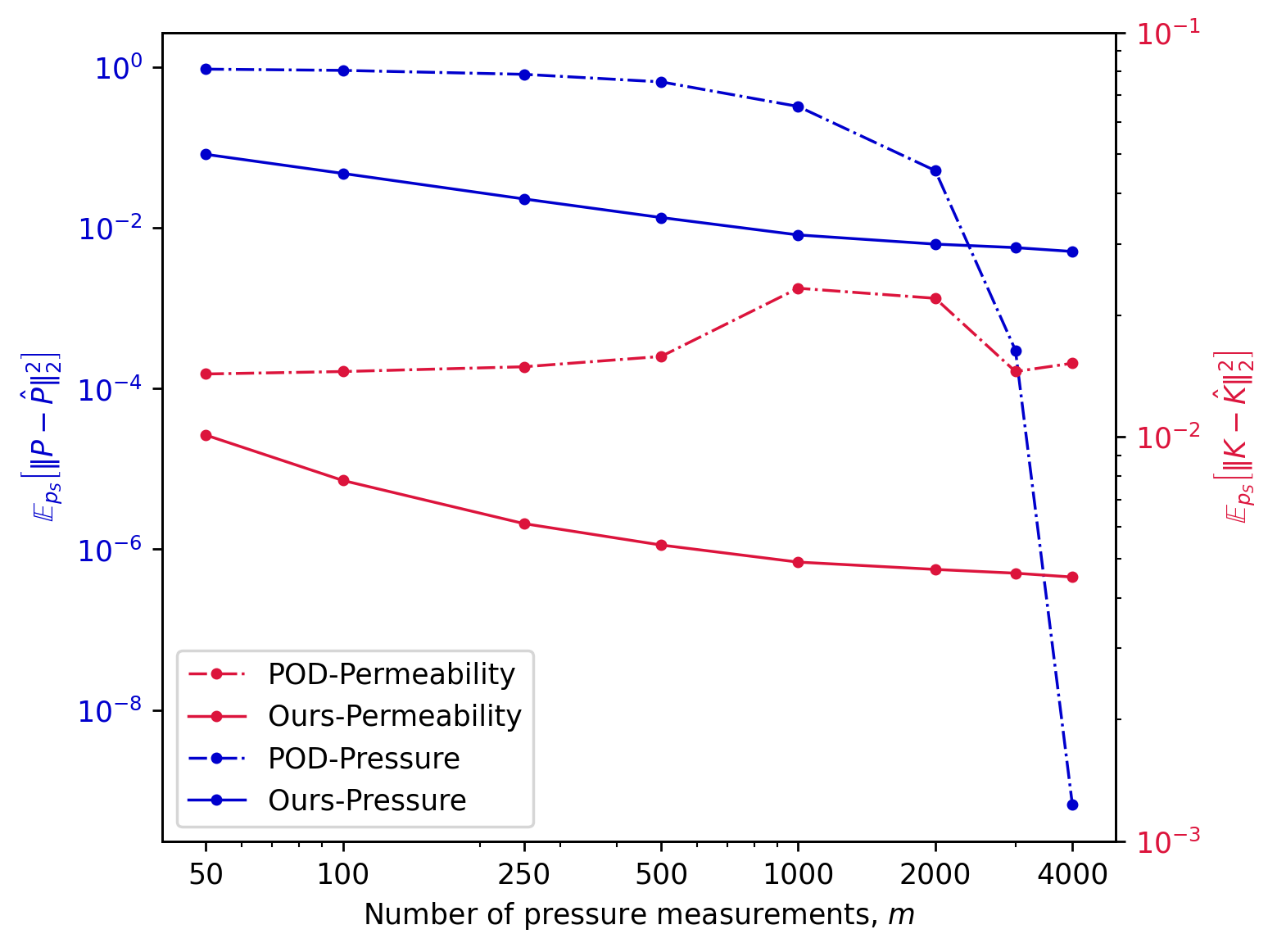}
        \caption{Reconstruction errors.}
        \label{fig:pod_recon_comparison}
    \end{subfigure}
    \begin{subfigure}[b]{0.45\textwidth}
        \includegraphics[width=\linewidth]{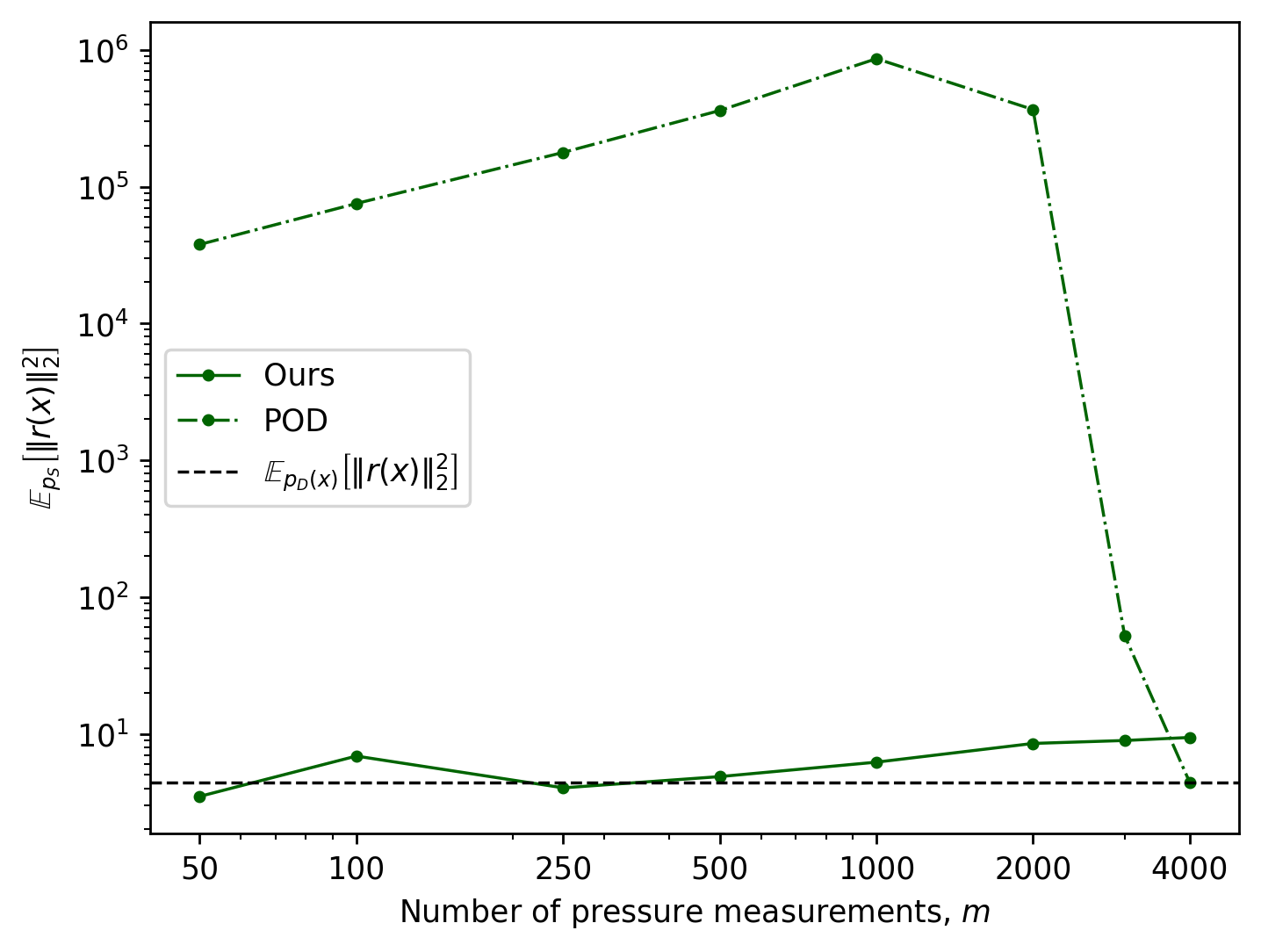}
        \caption{Physical residuals.}
        \label{fig:pod_residual_comparison}
    \end{subfigure}
    \caption{Our method greatly outperforms a POD-based method on field inversion and reconstruction from sparse measurements in terms of reconstruction error and physical residual. Sampling for our method is performed with physical consistency by solving the PF ODE with $\tau=2000$, $N=50$ and $M=10$.}
    \label{fig:pod_comparison}
\end{figure}

Solving the inverse problem using score-based generative models is further investigated by comparing it to a more established method of reconstructing data from measurements using a basis $\Psi$ obtained by performing proper orthogonal decomposition (POD) on the training dataset. Given a measurement matrix $P$ and partial field measurements $\mathbf{q}$, the full field is reconstructed by approximating the reconstruction in terms of the basis $\Psi$. This approximate reconstruction is computed as $\hat{\mathbf{y}} = \Psi[P\Psi]^\dagger\mathbf{q}$~\cite{chaturantabut2010nonlinear,wentland2023scalable}. The aim of comparison is to illustrate the effect that physical consistency sampling has on predictions when compared to another method. Here, we investigate the reconstructive performance of each method as a function of the number of pressure measurements taken, but also illustrate the average physical residual of reconstructions as a function of the number of measurements. Figure~\ref{fig:pod_recon_comparison} shows that for more sparsely sampled measurements, the score-based generative model is able to more accurately reconstruct both pressure and permeability fields. As there are no permeability measurements taken, the POD-based method cannot accurately reconstruct the permeability field, even in the limit of measuring the entire pressure field. However, this method does approach perfect reconstruction of the pressure field as the entire field is measured, while the score-based model does not. Regardless of the number of measurements taken, Fig.~\ref{fig:pod_residual_comparison} clearly shows that the score-based model with physical consistency sampling reliably produces samples which adhere to the physical PDE while the POD-based method produces predictions with orders of magnitude higher residuals. 

This particular example is intended to illustrate the potential for score-based generative models with physical consistency sampling in solving inverse problems, not to compare against advanced solutions. Whether the conditional distribution $p(x|\theta)$ has little uncertainty as in the forward problem demonstrated in Sec.~\ref{sec:surrogate_model} or is quite uncertain as in the case of few measurements here, generates samples will follow adhere to the underlying PDE.

%% file: 06_conclusion.tex
\section{Conclusions}\label{sec:conclusion}

Our work establishes a robust framework for developing score-based generative models that faithfully adhere to governing physical equations. Through the integration of physical consistency sampling and conditional score-based generative models, we present a versatile tool with potential for application to various tasks from the domain of physics. Key insights gleaned from our study underscore the importance of physical consistency sampling. Termed CoCoGen - signifying physical consistency and conditioning - this method consistently produces small residuals across diverse applications, alleviating the need for meticulous fine-tuning of architectures and hyperparameters for specific problems to achieve a similar effect. Leveraging the probability flow ordinary differential equation (PF ODE) contributes to smaller physical residuals over solver the reverse SDE, and adjusting the number of physical consistency steps $N$ not only reduces residuals of generated samples but also effectively mitigates noise, even on unaltered fields. However, proper hyperparameter tuning is still important, as an excessively small value for parameter $\tau$ can impede achieving small physical residuals even with physical consistency sampling. When appropriately configured, physical consistency sampling ensures that all generated samples align faithfully with the governing equations, irrespective of reconstruction error, preventing the model from predicting non-physical scenarios in a conditional or unconditional setting. We expect that more advanced methods of solving the reverse SDE or probability flow ODE may provide additional benefits in terms of minimizing residuals, especially when combined with physical consistency sampling. 

For the 1D Burgers equation, we considered time slabs of data and showed that low levels of error can be achieved compared to benchmark methods. For long time prediction,  the last time step can be used as the conditioning input and  unrolled predictions can be performed for future time steps.

The versatility of score-based generative models is evident in their capacity to address both forward and inverse problems. Physical consistency sampling further guarantees that all predictions are, at the very least, physically plausible.  Remarkably, this framework does not compromise prediction performance in terms of accuracy; instead, it systematically reduces physical residuals, affirming its efficacy over baseline sampling methods. In the Darcy flow example, the pressure field in the reverse process indirectly affects the permeability field via the score function estimation. This is shown to de-noise permeability fields and simultaneously lowers the physical residual. 

In essence, our work provides a flexible framework for leveraging score-based generative models in physics-based applications. Looking forward, we believe that the insights and methodologies presented herein pave the way for further exploration and application of score-based generative models in addressing several scientific machine learning tasks.

%% file: appendix.tex

\section{Solving Darcy Flow with a Linear System}\label{app:linear_system}

Section~\ref{sec:traditional_solve} in the main text describes the construction of the linear system $\mathbf{A}\mathbf{p} = \mathbf{f}$ to solve the Darcy flow equations for pressure using finite differences on a discretized computational domain. The matrix $\mathbf{A}$ is formed by constructing $n^2+1$ equations for pressure values at $n^2$ spatial locations in the computational domain. The equations are broken down into 4 categories: corner nodes, edge nodes, interior nodes, and integral enforcement.

There are four corner equations corresponding to nodes $1,1$, $n,1$, $1,n$, and $n,n$. These equations are given by
\begin{align*}
    \textrm{corner node } 1,1: & \; -\frac{K(\mathbf{x}_{1,1})}{\Delta x^2}[2p(\mathbf{x}_{2,1}) - 4p(\mathbf{x}_{1,1}) + 2p(\mathbf{x}_{1,2})] = f_s(\mathbf{x}_{1,1}) \\
    \textrm{corner node } n,1: & \; -\frac{K(\mathbf{x}_{n,1})}{\Delta x^2}[2p(\mathbf{x}_{n-1,1}) - 4p(\mathbf{x}_{n,1}) + 2p(\mathbf{x}_{n,2})] = f_s(\mathbf{x}_{n,1}) \\
    \textrm{corner node } 1,n: & \; -\frac{K(\mathbf{x}_{1,n})}{\Delta x^2}[2p(\mathbf{x}_{2,n}) - 4p(\mathbf{x}_{1,n}) + 2p(\mathbf{x}_{1,n-1})] = f_s(\mathbf{x}_{1,n}) \\
    \textrm{corner node } n,n: & \; -\frac{K(\mathbf{x}_{n,n})}{\Delta x^2}[2p(\mathbf{x}_{n-1,n}) - 4p(\mathbf{x}_{n,n}) + 2p(\mathbf{x}_{n,n-1})] = f_s(\mathbf{x}_{n,n}) 
\end{align*}

There are an additional $4(n-2)$ equations corresponding to nodes along edges which are not also corner nodes. The equations for each of these edges are given by
\begin{align*}
    \textrm{edge } i=1: \; &-\frac{K(\mathbf{x}_{1,j})}{\Delta x^2} [ p(\mathbf{x}_{1,j-1}) - 4p(\mathbf{x}_{1,j}) + p(\mathbf{x}_{1,j+1}) + 2p(\mathbf{x}_{2,j})] \\
    & - \frac{1}{2\Delta x}[K(\mathbf{x}_{1,j+1})-K(\mathbf{x}_{1,j-1})][p(\mathbf{x}_{1,j+1})-p(\mathbf{x}_{1,j-1})] = f_s(\mathbf{x}_{1,j})
\end{align*}
\begin{align*}
    \textrm{edge } i=n: \; &-\frac{K(\mathbf{x}_{n,j})}{\Delta x^2} [ p(\mathbf{x}_{n,j-1}) - 4p(\mathbf{x}_{n,j}) + p(\mathbf{x}_{n,j+1}) + 2p(\mathbf{x}_{n-1,j})] \\
    & - \frac{1}{2\Delta x}[K(\mathbf{x}_{n,j+1})-K(\mathbf{x}_{n,j-1})][p(\mathbf{x}_{n,j+1})-p(\mathbf{x}_{n,j-1})] = f_s(\mathbf{x}_{n,j})
\end{align*}
\begin{align*}
    \textrm{edge } j=1: \; &-\frac{K(\mathbf{x}_{i,1})}{\Delta x^2} [p(\mathbf{x}_{i-1,1}) - 4p(\mathbf{x}_{i,1}) + p(\mathbf{x}_{i+1,1}) + 2p(\mathbf{x}_{i,2})] \\
    & - \frac{1}{2\Delta x}[K(\mathbf{x}_{i+1,1})-K(\mathbf{x}_{i-1,1})][p(\mathbf{x}_{i+1,1})-p(\mathbf{x}_{i-1,1})] = f_s(\mathbf{x}_{i,1})
\end{align*}
\begin{align*}
    \textrm{edge } j=n: \; &-\frac{K(\mathbf{x}_{i,n})}{\Delta x^2} [p(\mathbf{x}_{i-1,n}) - 4p(\mathbf{x}_{i,n}) + p(\mathbf{x}_{i+1,n}) + 2p(\mathbf{x}_{i,n-1})] \\
    & - \frac{1}{2\Delta x}[K(\mathbf{x}_{i+1,n})-K(\mathbf{x}_{i-1,n})][p(\mathbf{x}_{i+1,n})-p(\mathbf{x}_{i-1,n})] = f_s(\mathbf{x}_{i,n})
\end{align*}
A remaining $(n-2)^2$ equations correspond to interior nodes (not on corners or edges) and are given by
\begin{align*}
    \textrm{interior node } i,j: \; &-\frac{K(\mathbf{x}_{i,j})}{\Delta x^2}[ p(\mathbf{x}_{i,j-1}+p(\mathbf{x}_{i-1,j}) - 4p(\mathbf{x}_{i,j}) + p(\mathbf{x}_{i+1,j}) + p(\mathbf{x}_{i,j+1})] \\
    & -\frac{1}{2\Delta x}[K(\mathbf{x}_{i+1,j}) - K(\mathbf{x}_{i-1,j})][p(\mathbf{x}_{i+1,j})-p(\mathbf{x}_{i-1,j})] \\
    & -\frac{1}{2\Delta x}[K(\mathbf{x}_{i,j+1}) - K(\mathbf{x}_{i,j-1})][p(\mathbf{x}_{i,j+1})-p(\mathbf{x}_{i,j-1})] = f_s(\mathbf{x}_{i,j})
\end{align*}

The final equation is formed by approximating the integral condition given in Eq.~\ref{eq:darcy} of the main text using 2-dimensional trapezoidal rule. This equation is given by
\begin{align*}
    \frac{\Delta x^2}{4} & \left [ p(\mathbf{x}_{1,1})+p(\mathbf{x}_{n,1})+p(\mathbf{x}_{1,n})+ p(\mathbf{x}_{n,n}) + 2\sum_{i=2}^{n-1} p(\mathbf{x}_{i,1}) + 2\sum_{i=2}^{n-1} p(\mathbf{x}_{i,n}) \right .  \\
    & \left . + 2\sum_{j=2}^{n-1} p(\mathbf{x}_{1,j}) + 2\sum_{j=2}^{n-1} p(\mathbf{x}_{n,j}) + 4\sum_{i=2}^{n-1}\sum_{j=2}^{n-1}p(\mathbf{x}_{i,j}) \right ] = 0
\end{align*}

We do not provide the explicit form of the matrix $\mathbf{A}$ due to the large form of the matrix, but it can be easily constructed from the provided equations. The overdetermined system is solved through least squares minimization where the solution is given by $\mathbf{p} = (\mathbf{A}^T\mathbf{A})^{-1}\mathbf{A}^T \mathbf{f}$.




\section{Unconditional Model and Training Details}\label{app:training_uncond}
Unconditional models for both the Darcy flow $s=16$ and $s=256$ datasets are identical in architecture and identical in training, containing 17.7M trainable parameters. We use a U-Net-type architecture consisting of convolution-based encoding and decoding layers described in the main text as the unconditional score-approximation model. This model is trained for 5000 epochs with a learning rate of $10^{-4}$ and batch size of 128. The SDE process which we employ is the variance-preserving SDE~\cite{song2021scorebased} with a linear $\beta(t)$ function corresponding to $\beta_{min}=10^{-4}$, $\beta_{max}=10$, and $T=1$. The optimization took 24.5 hours to complete 200,000 optimization iterations on 4 NVIDIA A6000 GPUs, which is sufficient to achieve convergence. This time includes sampling a batch of 8 fields every 1,000 optimization steps with each batch of samples taking an average of ~31 seconds to obtain. 

\section{Conditional Model and Training Details}
\subsection{Darcy flow and ControlNet}
\label{app:training_cond}
The conditional model architectures are exemplified by the ControlNet-based architecture in the main text. Each conditional augmentation has a slightly different architecture due to the need for the condition encoder to accept various forms in input. However, training is performed in an identical way to the unconditional training in~\ref{app:training_uncond}.

In the case of surrogate modeling, we use a condition encoder consisting of 3 linear layers followed by a reshaping operator and convolutional layers. The entire conditional augmentation in this case consists of 5.8M trainable parameters, with the 17.7M parameters in the unconditional model frozen during training. This results in a total of 23.5M parameters in the model. The optimization took 11.6 hours to achieve convergence for a total of 100,000 optimization iterations, less than half as long as training the unconditional model. 

The ControlNet architecture for performing field reconstruction and inversion is identical to that of the surrogate model ControlNet with the exception of the condition encoder. Here, the condition encoder consists of 3 convolutional layers only as the conditioning data is of the same shape as the input data. This results in a conditional augmentation with 5.7M trainable parameters, bringing the total parameters in the model to 23.4M including the frozen unconditional model. The optimization took 18.1 hours to achieve convergence for a total of 160,000 optimization iterations, a significant speedup over the unconditional model training.

\subsection{1D Burgers equation and EDM}
\label{app:edm}

The classifier-free guidance~\cite{ho2022classifier} demonstrates that the score function can be reformulated as follows when conditioning on the input $\theta$:

\begin{equation}
\label{eq:classifier_free_guidance}
\nabla_\mathbf{x} \log p_t(\mathbf{x}(t)|\theta)  = \gamma \underbrace{\nabla_x \log p_t(\mathbf{x}(t)|\theta)}_{conditional\, score} + (1-\gamma)\underbrace{\nabla_\mathbf{x} \log p_t(\mathbf{x}(t))}_{unconditional\, score}
\end{equation}
where $\gamma$ is the conditional scale. In the 1D Burgers equation, we set $\gamma=1$, and the original score function in the loss function is replaced by the conditional score function. For the 1D Burgers equation, the condition is the initial velocity profile, and the vector can be applied to the 2D channel via linear modulation with learnable coefficients.

To generate samples that condition on the initial velocity profile, we solve the PF ODE for 1000 steps. We introduce the physical consistency step from steps 505 to 1000, with a interval of 5 steps, and we vary the number of physical consistency steps after the PF ODE step from 0 to 5. The diffusion model has the same depth and width as the model used for the Darcy flow problem. We train for 1,000 epochs with a batch size of 32 and a learning rate of $1 \times 10^{-4}$ using a cosine annealing and warm-up schedule.

\section{Analytic Approximation to Conditional Sampling - Field Reconstruction and Inversion}\label{app:firms}


There exist a few special cases of conditional sampling in which it is possible to approximate conditional sampling after training only an unconditional generative model. Data imputation is one of such cases and attempts to predict or reconstruct unknown dimensions of data given some particular measured values of known dimensions. This is a conditional problem in which the aim is to sample from a conditional distribution in which conditioning information is the particular measured values $\omega$ of some known dimensions $\Omega(\mathbf{y}(0))$. The particular distribution to sample from is defined as $p(\mathbf{z}|\Omega(x(0))=\omega)$, where $\bar{\Omega}(\mathbf{y}(0)) = \mathbf{z}$ corresponds to the unknown dimensions. Given a trained unconditional score-based generative model, an approximation to this conditional sampling can be performed without any need for further training~\cite{song2021scorebased}. In our experiments, we measure only Darcy flow pressure fields at various spatial locations. From these measurements, the entire pressure field is reconstructed and an inverse problem is effectively solved to also approximate the permeability field. 
The examples in this work are intended to illustrate a framework which can be applied to any number of physical systems, effectively using measurements to predict physical fields which are consistent with governing equations. Although we perform this procedure with a closed form approximation based on a pretrained unconditional model, the technique can also be performed by training a conditional model as described in Sec.~\ref{sec:controlnet} of the main text. 

We perform data imputation by following the ideas from~\cite{song2021scorebased} in which $\Omega(\mathbf{y})$ denotes the known (measured) dimensions of $\mathbf{y}$ and $\bar{\Omega}(\mathbf{y})$ denotes the unknown dimensions of $\mathbf{y}$. Further, $f_{\bar{\Omega}}(\cdot, t)$ and $g_{\bar{\Omega}}(t)$ define a restriction of the operators $f(\cdot, t)$ and $g(t)$ in the SDE to the unknown dimensions only. 

As some portion of $\mathbf{y}$ is known, a new diffusion process is defined such that $\mathbf{z}(t) = \bar{\Omega}(\mathbf{y}(t))$ and
\[
d\mathbf{z} = f_{\bar{\Omega}}(\mathbf{z}, t)dt + g_{\bar{\Omega}}(t)dw \; .
\]
However, we do not desire to simply sample from $p(z(t=0))$, but rather $p(\mathbf{z}(t=0)|\Omega(\mathbf{y}(t=0))=\hat{\mathbf{y}})$, where $\hat{\mathbf{y}}$ are the measured values. The reverse process is therefore defined as
\[
d\mathbf{z} = [f_{\bar{\Omega}}(\mathbf{z}, t) - g_{\bar{\Omega}}(\mathbf{z}, t)^2\nabla_\mathbf{z} \log p_t(\mathbf{z}|\Omega(\mathbf{y}(0))=\hat{\mathbf{y}})]dt + g_{\bar{\Omega}}(t)dw
\]
It is shown in~\cite{song2021scorebased} that $\nabla_\mathbf{z} \log p_t(\mathbf{z}|\Omega(\mathbf{y}(0))=\hat{\mathbf{y}})$ can be approximated by $\nabla_\mathbf{z} \log p_t(\mathbf{u}(t))$ where $\mathbf{u}(t) = [\mathbf{z}(t); \hat{\Omega}(\mathbf{y}(t))]$ is a vector such that $\Omega(\mathbf{u}(t)) = \hat{\Omega}(\mathbf{y}(t))$ and $\bar{\Omega}(\mathbf{u}(t)) = \mathbf{z}(t)$. The quantity $\hat{\Omega}(\mathbf{y}(t))$ denotes a random sample from $p_t(\Omega(\mathbf{y}(t))|\Omega(\mathbf{y}(0))=\hat{\mathbf{y}})$, which can typically be obtained by sampling the corresponding known dimensions from the forward SDE process as the dimensions of $\mathbf{y}$ are uncorrelated in the forward process (Eq.~\ref{eq:training_loss} of the main text). The vector $\mathbf{u}(t)$ contains all known and unknown components of $\mathbf{y}$ and therefore $s_\theta(\mathbf{y}(t), t) \approx \nabla_\mathbf{z} \log p_t(\mathbf{u}(t)) \approx \nabla_\mathbf{z} \log p_t(\mathbf{z}|\Omega(\mathbf{y}(0))=\hat{\mathbf{y}})$. Thus, an approximation to $\nabla_\mathbf{z} \log p_t(\mathbf{z}|\Omega(\mathbf{y}(0))=\hat{\mathbf{y}})$ can be computed by leveraging the score function approximation $s_\theta(\mathbf{y}(t), t)$ without any additional special training by sampling the known dimensions of $\mathbf{y}(t)$ at each time step according to the forward process. Further, sampling with physical consistency steps (Sec.~\ref{sec:physical_consistency} of the main text) can still be performed in the same manner, leading to samples which are consistent with the governing PDE. 

\begin{table}[!htb]
\centering
\caption{Field reconstruction and inversion (no RePaint) with various number of pressure measurements (Darcy Flow, $s=16$). Analytic approximation to data imputation used in sampling.}
\label{tab:reconstruction_darcy_n16}
\begin{tabularx}{\textwidth}{cccccc}
\hline
Equation & $m$ & $\mathbb{E}_{p_S(\mathbf{y})}[||\mathbf{r}||_2^2]$ & $\mathbb{E}_{p_S(\mathbf{y})}[||\mathbf{r}||_2^2] - \mathbb{E}_{p_D(\mathbf{y})}[||\mathbf{r}||_2^2]$ & $\mathbb{E}_{p_D(\mathbf{y})}[||P - \hat{P}||_2^2]$ & $\mathbb{E}_{p_D(\mathbf{y})}[||K - \hat{K}||_2^2]$\\
\hline
SDE & 0 & 3.6 & -0.9 & $1.02\times 10^{-1}$ & $8.8\times 10^{-3}$\\
\hline
SDE & 10 & 224.1 & 219.6 & $1.04\times 10^{-1}$ & $9.5\times 10^{-3}$\\
\hline
SDE & 50 & 1463.5 & 1459.0 & $8.73\times 10^{-2}$ & $9.1\times 10^{-3}$\\
\hline
SDE & 100 & 3986.1 & 3981.6 & $5.92\times 10^{-2}$ & $8.5\times 10^{-3}$\\
\hline
SDE & 250 & 4142.1 & 4137.6 & $3.25\times 10^{-2}$ & $7.8\times 10^{-3}$\\
\hline
SDE & 500 & 1970.2 & 1965.7 & $9.1\times 10^{-3}$ & $6.4\times 10^{-3}$\\
\hline
SDE & 1000 & 183.8 & 179.3 & $5.0\times 10^{-4}$ & $4.3\times 10^{-3}$\\
\hline
SDE & 2000 & 7.5 & 3.0 & $3.6\times 10^{-6}$ & $2.8\times 10^{-3}$\\
\hline
SDE & 4000 & 7.6 & 3.1 & $2.4\times 10^{-7}$ & $2.0\times 10^{-3}$\\
\hline

\end{tabularx}
\end{table}

It is worth mentioning that performing data imputation in this special case can be orders of magnitude more expensive when solving the PF ODE reverse process rather than the reverse SDE process. It is illustrated in Sec.~\ref{sec:physical_consistency} of the main text that solving the PF ODE reverse process results in improved physical residual minimization with smaller PDE operator evaluations when compared to the reverse SDE. However, the outlined method of data imputation requires sampling from the forward process to obtain $\hat{\Omega}(\mathbf{y}(t))$. Sampling in time via the forward PF ODE process involves solving the PF ODE forward in time. In contrast, the forward SDE has a closed form solution to obtain samples at any future time (Eq.~\ref{eq:transition_kernel} in the main text), rendering data imputation (field reconstruction and inversion) far more feasible and inexpensive.

\begin{table}[!htb]
\centering
\caption{Field reconstruction and inversion with RePaint performance with the number of repainting steps $r$ (Darcy Flow, $s=16$, $m=500$).}
\label{tab:repaint_darcy_n16}
\begin{tabularx}{\textwidth}{cccccc}
\hline
Equation & $r$ & $\mathbb{E}_{p_S(\mathbf{y})}[||\mathbf{r}||_2^2]$ & $\mathbb{E}_{p_S(\mathbf{y})}[||\mathbf{r}||_2^2] - \mathbb{E}_{p_D(\mathbf{y})}[||\mathbf{r}||_2^2]$ & $\mathbb{E}_{p_D(\mathbf{y})}[||P - \hat{P}||_2^2]$ & $\mathbb{E}_{p_D(\mathbf{y})}[||K - \hat{K}||_2^2]$\\
\hline
SDE & 1 & 750.7 & 746.2 & $4.9\times 10^{-3}$ & $4.3\times 10^{-3}$\\
\hline
SDE & 2 & 26.2 & 21.7 & $3.0\times 10^{-4}$ & $2.9\times 10^{-3}$\\
\hline
SDE & 3 & 8.7 & 4.2 & $3.1\times 10^{-5}$ & $3.8\times 10^{-3}$\\
\hline
SDE & 5 & 10.6 & 6.1 & $1.9\times 10^{-5}$ & $2.9\times 10^{-3}$\\
\hline
SDE & 10 & 9.6 & 5.1 & $1.7\times 10^{-5}$ & $4.8\times 10^{-3}$\\
\hline
\end{tabularx}
\end{table}

We use the same score-based generative model defined in the experiments of Sec.~\ref{sec:physical_consistency} in the main text in our experiments. However, only the reverse SDE is solved and not the PF ODE due to aforementioned inefficiencies. We perform the baseline data imputation approximation for various number of measured dimensions. The number of measurements is defined as $m$. To perform data imputation, we randomly select $m$ spatial locations to measure only pressure in the physical domain, similar to Sec.~\ref{sec:firms} in the main text. The random seed is kept constant such that for two cases in which $m_2 > m_1$, the spatial locations of case 1 are a subset of the spatial locations of case 2. 

\begin{figure}[h!]
    \centering
    \begin{subfigure}[b]{0.49\textwidth}
        \includegraphics[width=\linewidth]{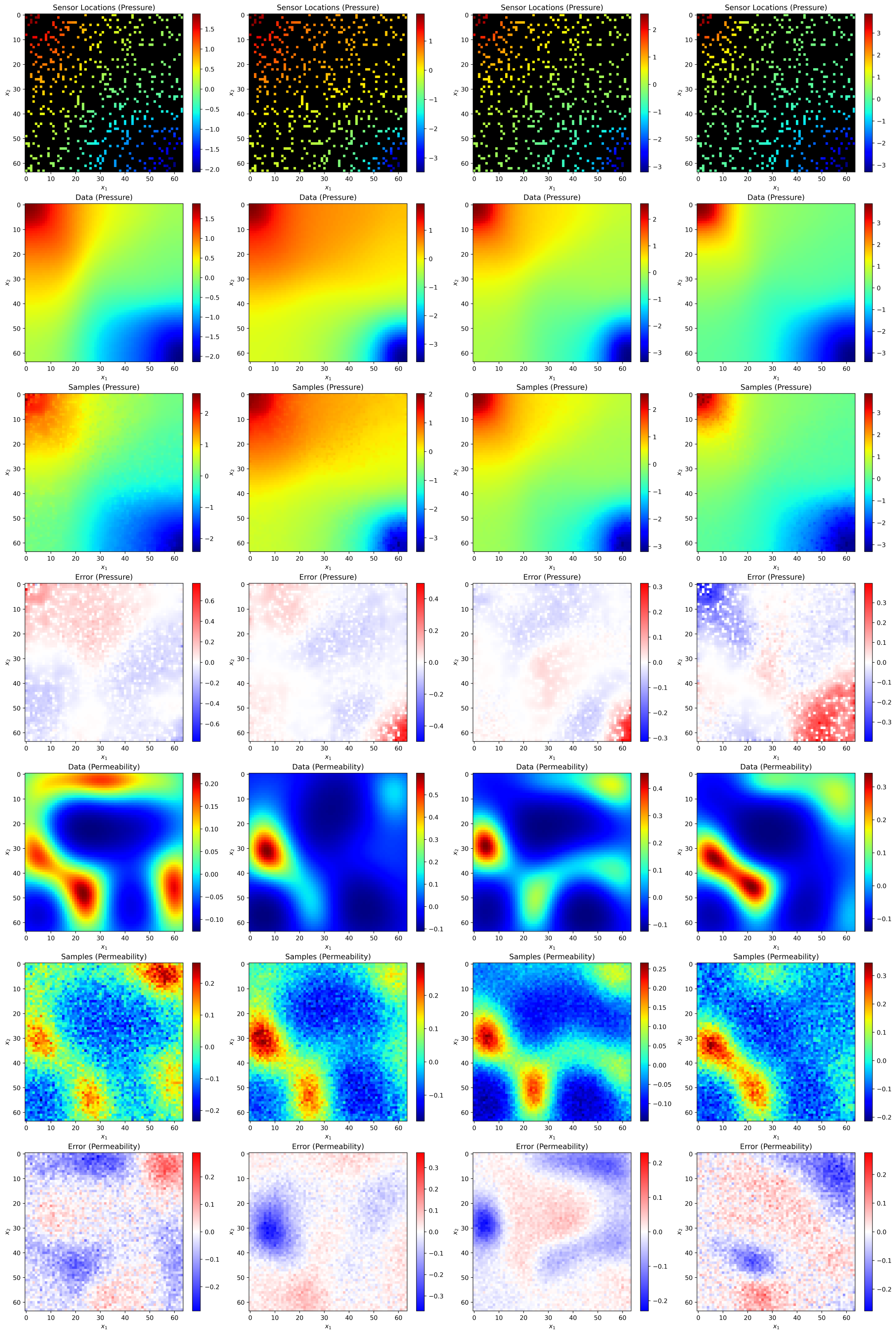}
        \caption{$r=0$}
        \label{subfig:uncond_firms_no_repaint}
    \end{subfigure}
    \begin{subfigure}[b]{0.49\textwidth}
        \includegraphics[width=\linewidth]{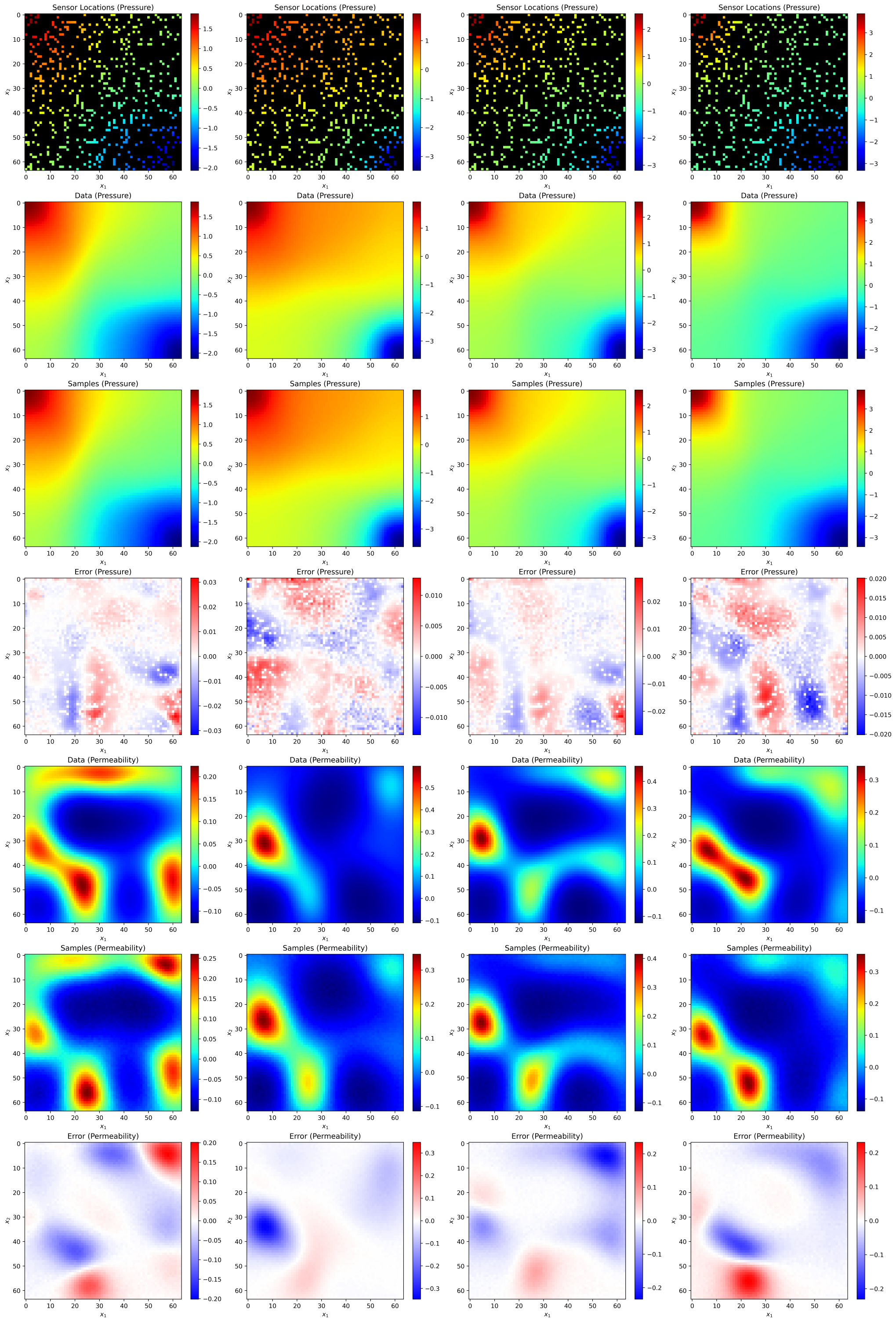}
        \caption{$r=5$}
        \label{subfig:uncond_firms_repaint}
    \end{subfigure}
    \caption{Field reconstruction and inversion (a) without and (b) with RePaint. RePaint incorporates more information about the known dimensions into the unknown dimensions. Figure is high quality, zoom in for more clarity.}
    \label{fig:reconstruction_darcy_base_n16}
\end{figure}

For each experiment, we vary only the number of pressure measurements used to reconstruct the pressure field and invert the permeability field. The reverse process uses an Euler-Maruyama time integration method to solve the reverse SDE with $\tau = 2,000$ time steps. We also use $N=50$ physical consistency steps and $M=10$ additional physical consistency steps. For these experiments, we illustrate results on the Darcy flow $s=16$ dataset only.


Pressure data is denoted $P$ with reconstructions denoted $\hat{P}$. Similarly, permeability data and reconstructions are denoted $K$ and $\hat{K}$ respectively. Table~\ref{tab:reconstruction_darcy_n16} 
gives quantitative reconstruction results, and Fig.~\ref{subfig:uncond_firms_no_repaint} illustrates qualitative reconstruction on the test set.
Although the reconstruction for both pressure and permeability fields improves with increasing number of samples, the physical residual is very high, particularly for smaller number of measurements, even with physical consistency enforcement. This is likely due to the incoherence between the known (measured) and unknown regions of the data samples. It is evident in Fig.~\ref{fig:reconstruction_darcy_base_n16} that the measured points do not match well with the reconstructed points. 

To reduce this discrepancy, we turn towards a recent work addressing this issue called RePaint~\cite{repaint}. This effectively alters between reverse and forward steps to reduce the boundary discrepancy between measured and unmeasured regions of the data samples. At each discrete time step $i$ of the reverse SDE solver, an additional $r$ \emph{resampling} steps are performed. A resampling step consists of solving the forward SDE for a single time step, followed by solving the reverse SDE for the same time step, effectively alternating $r$ times between stepping forward and stepping backward in time according to the SDE. This allows better mixing of the measured and unmeasured dimensions of the data, adding artificial noise before denoising, effectively incorporating more information about the known dimensions to the unknown dimensions at each resampling step. However, this comes at a significant cost: the number of score function evaluations scales linearly with $r$. This can be partially alleviated by performing $r$ resampling steps only at some discrete pre-selected number of time steps instead of at each time step. However, this presents additional difficulties such as determining the optimal time steps to perform resampling at. In our experiments, we perform resampling $r$ times at each time step.


We investigate the power of RePainting by setting a constant number of pressure measurements ($m=500$), and solving the reverse SDE with various numbers of resampling steps. Table~\ref{tab:repaint_darcy_n16} illustrates the results of this experiment, showing that the resampling procedure greatly improves reconstruction performance and drastically reduces physical residuals over the standard method of sampling without RePaint ($r=0$). Additionally, reconstructions rapidly improve with increasing number of resampling steps. Sampling with RePaint and $r=3$ reduces the average physical residual from $1970.2$ to $8.7$ over sampling without RePaint. The average reconstruction error is also reduced from $9.1\times 10^{-3}$ to $3.1\times 10^{-5}$ and the field inversion error is reduced from $6.4\times 10^{-3}$ to $3.8\times 10^{-3}$. Figure~\ref{subfig:uncond_firms_repaint} qualitatively shows the effect that RePaint has on sample quality. When compared to sampling without RePainting (Fig.~\ref{subfig:uncond_firms_no_repaint}), reconstructions are far more accurate and predictions less noisy, especially permeability fields. This is a direct result of reducing the boundary discrepancy using the RePainting algorithm, leading to de-noised samples.

Performing field reconstruction and inversion as described in this section outperforms the method of conditional generation using ControlNet detailed in the main text in terms of reconstruction performance. However, using the repainting algorithm linearly increases the cost of sampling as a function or $r$. Thus, the best results here require around 5x more compute time to obtain over those obtained using ControlNet-based augmentations for conditional generation.

%% file: main.bbl
\begin{thebibliography}{10}
\expandafter\ifx\csname url\endcsname\relax
  \def\url#1{\texttt{#1}}\fi
\expandafter\ifx\csname urlprefix\endcsname\relax\def\urlprefix{URL }\fi
\expandafter\ifx\csname href\endcsname\relax
  \def\href#1#2{#2} \def\path#1{#1}\fi

\bibitem{kingma2014autoencoding}
D.~Kingma, M.~Welling, {Auto-Encoding Variational Bayes}, International Conference on Learning Representations (12 2013).

\bibitem{pmlr-v32-rezende14}
D.~J. Rezende, S.~Mohamed, D.~Wierstra, \href{https://proceedings.mlr.press/v32/rezende14.html}{Stochastic backpropagation and approximate inference in deep generative models}, in: E.~P. Xing, T.~Jebara (Eds.), Proceedings of the 31st International Conference on Machine Learning, Vol.~32 of Proceedings of Machine Learning Research, PMLR, Bejing, China, 2014, pp. 1278--1286.
\newline\urlprefix\url{https://proceedings.mlr.press/v32/rezende14.html}

\bibitem{Jacobsen2022}
C.~Jacobsen, K.~Duraisamy, \href{https://www.frontiersin.org/articles/10.3389/fphy.2022.890910}{Disentangling generative factors of physical fields using variational autoencoders}, Frontiers in Physics 10 (2022).
\newblock \href {https://doi.org/10.3389/fphy.2022.890910} {\path{doi:10.3389/fphy.2022.890910}}.
\newline\urlprefix\url{https://www.frontiersin.org/articles/10.3389/fphy.2022.890910}

\bibitem{pmlr-v15-larochelle11a}
H.~Larochelle, I.~Murray, \href{https://proceedings.mlr.press/v15/larochelle11a.html}{The neural autoregressive distribution estimator}, in: G.~Gordon, D.~Dunson, M.~Dudík (Eds.), Proceedings of the Fourteenth International Conference on Artificial Intelligence and Statistics, Vol.~15 of Proceedings of Machine Learning Research, PMLR, Fort Lauderdale, FL, USA, 2011, pp. 29--37.
\newline\urlprefix\url{https://proceedings.mlr.press/v15/larochelle11a.html}

\bibitem{pmlr-v37-germain15}
M.~Germain, K.~Gregor, I.~Murray, H.~Larochelle, \href{https://proceedings.mlr.press/v37/germain15.html}{Made: Masked autoencoder for distribution estimation}, in: F.~Bach, D.~Blei (Eds.), Proceedings of the 32nd International Conference on Machine Learning, Vol.~37 of Proceedings of Machine Learning Research, PMLR, Lille, France, 2015, pp. 881--889.
\newline\urlprefix\url{https://proceedings.mlr.press/v37/germain15.html}

\bibitem{pmlr-v48-oord16}
A.~van~den Oord, N.~Kalchbrenner, K.~Kavukcuoglu, \href{https://proceedings.mlr.press/v48/oord16.html}{Pixel recurrent neural networks}, in: M.~F. Balcan, K.~Q. Weinberger (Eds.), Proceedings of The 33rd International Conference on Machine Learning, Vol.~48 of Proceedings of Machine Learning Research, PMLR, New York, New York, USA, 2016, pp. 1747--1756.
\newline\urlprefix\url{https://proceedings.mlr.press/v48/oord16.html}

\bibitem{ebm1}
Y.~Lecun, S.~Chopra, R.~Hadsell, M.~Ranzato, F.~Huang, A tutorial on energy-based learning, MIT Press, Cambridge, 2006.

\bibitem{song2021train}
Y.~Song, D.~P. Kingma, \href{https://arxiv.org/abs/2101.03288}{How to train your energy-based models} (2021).
\newblock \href {http://arxiv.org/abs/2101.03288} {\path{arXiv:2101.03288}}.
\newline\urlprefix\url{https://arxiv.org/abs/2101.03288}

\bibitem{pmlr-v37-rezende15}
D.~Rezende, S.~Mohamed, \href{https://proceedings.mlr.press/v37/rezende15.html}{Variational inference with normalizing flows}, in: F.~Bach, D.~Blei (Eds.), Proceedings of the 32nd International Conference on Machine Learning, Vol.~37 of Proceedings of Machine Learning Research, PMLR, Lille, France, 2015, pp. 1530--1538.
\newline\urlprefix\url{https://proceedings.mlr.press/v37/rezende15.html}

\bibitem{dinh2015nice}
L.~Dinh, D.~Krueger, Y.~Bengio, \href{https://arxiv.org/abs/1410.8516v6}{Nice: Non-linear independent components estimation} (2015).
\newblock \href {http://arxiv.org/abs/1410.8516} {\path{arXiv:1410.8516}}.
\newline\urlprefix\url{https://arxiv.org/abs/1410.8516v6}

\bibitem{dinh2017density}
L.~Dinh, J.~Sohl-Dickstein, S.~Bengio, \href{https://openreview.net/forum?id=HkpbnH9lx}{Density estimation using real {NVP}}, in: International Conference on Learning Representations, 2017.
\newline\urlprefix\url{https://openreview.net/forum?id=HkpbnH9lx}

\bibitem{NIPS2014_5ca3e9b1}
I.~Goodfellow, J.~Pouget-Abadie, M.~Mirza, B.~Xu, D.~Warde-Farley, S.~Ozair, A.~Courville, Y.~Bengio, \href{https://proceedings.neurips.cc/paper_files/paper/2014/file/5ca3e9b122f61f8f06494c97b1afccf3-Paper.pdf}{Generative adversarial nets}, in: Z.~Ghahramani, M.~Welling, C.~Cortes, N.~Lawrence, K.~Weinberger (Eds.), Advances in Neural Information Processing Systems, Vol.~27, 2014.
\newline\urlprefix\url{https://proceedings.neurips.cc/paper_files/paper/2014/file/5ca3e9b122f61f8f06494c97b1afccf3-Paper.pdf}

\bibitem{pmlr-v48-chwialkowski16}
K.~Chwialkowski, H.~Strathmann, A.~Gretton, \href{https://proceedings.mlr.press/v48/chwialkowski16.html}{A kernel test of goodness of fit}, in: M.~F. Balcan, K.~Q. Weinberger (Eds.), Proceedings of The 33rd International Conference on Machine Learning, Vol.~48 of Proceedings of Machine Learning Research, PMLR, New York, New York, USA, 2016, pp. 2606--2615.
\newline\urlprefix\url{https://proceedings.mlr.press/v48/chwialkowski16.html}

\bibitem{song2021scorebased}
Y.~Song, J.~Sohl-Dickstein, D.~P. Kingma, A.~Kumar, S.~Ermon, B.~Poole, \href{https://openreview.net/forum?id=PxTIG12RRHS}{Score-based generative modeling through stochastic differential equations}, in: International Conference on Learning Representations, 2021.
\newline\urlprefix\url{https://openreview.net/forum?id=PxTIG12RRHS}

\bibitem{NEURIPS2020_4c5bcfec}
J.~Ho, A.~Jain, P.~Abbeel, \href{https://proceedings.neurips.cc/paper_files/paper/2020/file/4c5bcfec8584af0d967f1ab10179ca4b-Paper.pdf}{Denoising diffusion probabilistic models}, in: H.~Larochelle, M.~Ranzato, R.~Hadsell, M.~Balcan, H.~Lin (Eds.), Advances in Neural Information Processing Systems, Vol.~33, 2020, pp. 6840--6851.
\newline\urlprefix\url{https://proceedings.neurips.cc/paper_files/paper/2020/file/4c5bcfec8584af0d967f1ab10179ca4b-Paper.pdf}

\bibitem{NEURIPS2019_3001ef25}
Y.~Song, S.~Ermon, \href{https://proceedings.neurips.cc/paper_files/paper/2019/file/3001ef257407d5a371a96dcd947c7d93-Paper.pdf}{Generative modeling by estimating gradients of the data distribution}, in: H.~Wallach, H.~Larochelle, A.~Beygelzimer, F.~d\textquotesingle Alch\'{e}-Buc, E.~Fox, R.~Garnett (Eds.), Advances in Neural Information Processing Systems, Vol.~32, 2019.
\newline\urlprefix\url{https://proceedings.neurips.cc/paper_files/paper/2019/file/3001ef257407d5a371a96dcd947c7d93-Paper.pdf}

\bibitem{xu2022poisson}
Y.~Xu, Z.~Liu, M.~Tegmark, T.~S. Jaakkola, \href{https://openreview.net/forum?id=voV_TRqcWh}{Poisson flow generative models}, in: A.~H. Oh, A.~Agarwal, D.~Belgrave, K.~Cho (Eds.), Advances in Neural Information Processing Systems, 2022.
\newline\urlprefix\url{https://openreview.net/forum?id=voV_TRqcWh}

\bibitem{xu2023pfgm}
Y.~Xu, Z.~Liu, Y.~Tian, S.~Tong, M.~Tegmark, T.~Jaakkola, \href{https://arxiv.org/abs/2302.04265}{Pfgm++: Unlocking the potential of physics-inspired generative models} (2023).
\newblock \href {http://arxiv.org/abs/2302.04265} {\path{arXiv:2302.04265}}.
\newline\urlprefix\url{https://arxiv.org/abs/2302.04265}

\bibitem{kim2023consistency}
D.~Kim, C.-H. Lai, W.-H. Liao, N.~Murata, Y.~Takida, T.~Uesaka, Y.~He, Y.~Mitsufuji, S.~Ermon, \href{https://arxiv.org/abs/2310.02279}{Consistency trajectory models: Learning probability flow ode trajectory of diffusion} (2023).
\newblock \href {http://arxiv.org/abs/2310.02279} {\path{arXiv:2310.02279}}.
\newline\urlprefix\url{https://arxiv.org/abs/2310.02279}

\bibitem{austin2021structured}
J.~Austin, D.~D. Johnson, J.~Ho, D.~Tarlow, R.~van~den Berg, \href{https://openreview.net/forum?id=h7-XixPCAL}{Structured denoising diffusion models in discrete state-spaces}, in: A.~Beygelzimer, Y.~Dauphin, P.~Liang, J.~W. Vaughan (Eds.), Advances in Neural Information Processing Systems, 2021.
\newline\urlprefix\url{https://openreview.net/forum?id=h7-XixPCAL}

\bibitem{li2022diffusionlm}
X.~L. Li, J.~Thickstun, I.~Gulrajani, P.~Liang, T.~Hashimoto, \href{https://openreview.net/forum?id=3s9IrEsjLyk}{Diffusion-{LM} improves controllable text generation}, in: A.~H. Oh, A.~Agarwal, D.~Belgrave, K.~Cho (Eds.), Advances in Neural Information Processing Systems, 2022.
\newline\urlprefix\url{https://openreview.net/forum?id=3s9IrEsjLyk}

\bibitem{gong2023diffuseq}
S.~Gong, M.~Li, J.~Feng, Z.~Wu, L.~Kong, \href{https://openreview.net/forum?id=jQj-_rLVXsj}{Diffuseq: Sequence to sequence text generation with diffusion models}, in: The Eleventh International Conference on Learning Representations, 2023.
\newline\urlprefix\url{https://openreview.net/forum?id=jQj-_rLVXsj}

\bibitem{osti_10433150}
X.~Han, S.~Kumar, Y.~Tsvetkov, \href{https://par.nsf.gov/biblio/10433150}{Ssd-lm: Semi-autoregressive simplex-based diffusion language model for text generation and modular control}, ACL: Annual Meeting of the Association for Computational Linguistics.
\newline\urlprefix\url{https://par.nsf.gov/biblio/10433150}

\bibitem{9887996}
C.~Saharia, J.~Ho, W.~Chan, T.~Salimans, D.~J. Fleet, M.~Norouzi, Image super-resolution via iterative refinement, IEEE Transactions on Pattern Analysis and Machine Intelligence 45~(4) (2023) 4713--4726.
\newblock \href {https://doi.org/10.1109/TPAMI.2022.3204461} {\path{doi:10.1109/TPAMI.2022.3204461}}.

\bibitem{repaint}
A.~Lugmayr, M.~Danelljan, A.~Romero, F.~Yu, R.~Timofte, L.~Van~Gool, Repaint: Inpainting using denoising diffusion probabilistic models, in: 2022 IEEE/CVF Conference on Computer Vision and Pattern Recognition (CVPR), 2022, pp. 11451--11461.
\newblock \href {https://doi.org/10.1109/CVPR52688.2022.01117} {\path{doi:10.1109/CVPR52688.2022.01117}}.

\bibitem{10.1145/3528233.3530757}
C.~Saharia, W.~Chan, H.~Chang, C.~Lee, J.~Ho, T.~Salimans, D.~Fleet, M.~Norouzi, \href{https://doi.org/10.1145/3528233.3530757}{Palette: Image-to-image diffusion models}, in: ACM SIGGRAPH 2022 Conference Proceedings, SIGGRAPH '22, Association for Computing Machinery, New York, NY, USA, 2022.
\newblock \href {https://doi.org/10.1145/3528233.3530757} {\path{doi:10.1145/3528233.3530757}}.
\newline\urlprefix\url{https://doi.org/10.1145/3528233.3530757}

\bibitem{10203350}
J.~Xu, S.~Liu, A.~Vahdat, W.~Byeon, X.~Wang, S.~D. Mello, \href{https://doi.ieeecomputersociety.org/10.1109/CVPR52729.2023.00289}{Open-vocabulary panoptic segmentation with text-to-image diffusion models}, in: 2023 IEEE/CVF Conference on Computer Vision and Pattern Recognition (CVPR), IEEE Computer Society, Los Alamitos, CA, USA, 2023, pp. 2955--2966.
\newblock \href {https://doi.org/10.1109/CVPR52729.2023.00289} {\path{doi:10.1109/CVPR52729.2023.00289}}.
\newline\urlprefix\url{https://doi.ieeecomputersociety.org/10.1109/CVPR52729.2023.00289}

\bibitem{rombach2022high}
R.~Rombach, A.~Blattmann, D.~Lorenz, P.~Esser, B.~Ommer, High-resolution image synthesis with latent diffusion models, in: Proceedings of the IEEE/CVF Conference on Computer Vision and Pattern Recognition, 2022, pp. 10684--10695.

\bibitem{ramesh2022hierarchical}
A.~Ramesh, P.~Dhariwal, A.~Nichol, C.~Chu, M.~Chen, \href{https://arxiv.org/abs/2204.06125}{Hierarchical text-conditional image generation with clip latents} (2022).
\newblock \href {http://arxiv.org/abs/2204.06125} {\path{arXiv:2204.06125}}.
\newline\urlprefix\url{https://arxiv.org/abs/2204.06125}

\bibitem{nichol2022glide}
A.~Nichol, P.~Dhariwal, A.~Ramesh, P.~Shyam, P.~Mishkin, B.~McGrew, I.~Sutskever, M.~Chen, Glide: Towards photorealistic image generation and editing with text-guided diffusion models, in: International Conference on Machine Learning, 2022.

\bibitem{ruiz2022dreambooth}
N.~Ruiz, Y.~Li, V.~Jampani, Y.~Pritch, M.~Rubinstein, K.~Aberman, \href{https://arxiv.org/abs/2208.12242}{Dreambooth: Fine tuning text-to-image diffusion models for subject-driven generation} (2022).
\newline\urlprefix\url{https://arxiv.org/abs/2208.12242}

\bibitem{controlnet}
M.~A. Lvmin~Zhang, Anyi~Rao, Adding conditional control to text-to-image diffusion models, in: International Conference on Computer Vision, 2023.

\bibitem{guan2023d}
J.~Guan, W.~W. Qian, X.~Peng, Y.~Su, J.~Peng, J.~Ma, \href{https://openreview.net/forum?id=kJqXEPXMsE0}{3d equivariant diffusion for target-aware molecule generation and affinity prediction}, in: The Eleventh International Conference on Learning Representations, 2023.
\newline\urlprefix\url{https://openreview.net/forum?id=kJqXEPXMsE0}

\bibitem{luo2022antigenspecific}
S.~Luo, Y.~Su, X.~Peng, S.~Wang, J.~Peng, J.~Ma, \href{https://openreview.net/forum?id=jSorGn2Tjg}{Antigen-specific antibody design and optimization with diffusion-based generative models for protein structures}, in: A.~H. Oh, A.~Agarwal, D.~Belgrave, K.~Cho (Eds.), Advances in Neural Information Processing Systems, 2022.
\newline\urlprefix\url{https://openreview.net/forum?id=jSorGn2Tjg}

\bibitem{song2022solving}
Y.~Song, L.~Shen, L.~Xing, S.~Ermon, \href{https://openreview.net/forum?id=vaRCHVj0uGI}{Solving inverse problems in medical imaging with score-based generative models}, in: International Conference on Learning Representations, 2022.
\newline\urlprefix\url{https://openreview.net/forum?id=vaRCHVj0uGI}

\bibitem{CHUNG2022102479}
H.~Chung, J.~C. Ye, \href{https://www.sciencedirect.com/science/article/pii/S1361841522001268}{Score-based diffusion models for accelerated mri}, Medical Image Analysis 80 (2022) 102479.
\newblock \href {https://doi.org/https://doi.org/10.1016/j.media.2022.102479} {\path{doi:https://doi.org/10.1016/j.media.2022.102479}}.
\newline\urlprefix\url{https://www.sciencedirect.com/science/article/pii/S1361841522001268}

\bibitem{Krizhevsky2009LearningML}
A.~Krizhevsky, \href{https://api.semanticscholar.org/CorpusID:18268744}{Learning multiple layers of features from tiny images}, 2009.
\newline\urlprefix\url{https://api.semanticscholar.org/CorpusID:18268744}

\bibitem{10.1145/3626235}
L.~Yang, Z.~Zhang, Y.~Song, S.~Hong, R.~Xu, Y.~Zhao, W.~Zhang, B.~Cui, M.-H. Yang, \href{https://doi.org/10.1145/3626235}{Diffusion models: A comprehensive survey of methods and applications}, ACM Comput. Surv. 56~(4) (nov 2023).
\newblock \href {https://doi.org/10.1145/3626235} {\path{doi:10.1145/3626235}}.
\newline\urlprefix\url{https://doi.org/10.1145/3626235}

\bibitem{yang2023denoising}
G.~Yang, S.~Sommer, \href{https://arxiv.org/abs/2301.11661}{A denoising diffusion model for fluid field prediction} (2023).
\newblock \href {http://arxiv.org/abs/2301.11661} {\path{arXiv:2301.11661}}.
\newline\urlprefix\url{https://arxiv.org/abs/2301.11661}

\bibitem{10.1007/978-3-319-24574-4_28}
O.~Ronneberger, P.~Fischer, T.~Brox, U-net: Convolutional networks for biomedical image segmentation, in: N.~Navab, J.~Hornegger, W.~M. Wells, A.~F. Frangi (Eds.), Medical Image Computing and Computer-Assisted Intervention -- MICCAI 2015, Springer International Publishing, Cham, 2015, pp. 234--241.

\bibitem{RAISSI2019686}
M.~Raissi, P.~Perdikaris, G.~Karniadakis, \href{https://www.sciencedirect.com/science/article/pii/S0021999118307125}{Physics-informed neural networks: A deep learning framework for solving forward and inverse problems involving nonlinear partial differential equations}, Journal of Computational Physics 378 (2019) 686--707.
\newblock \href {https://doi.org/https://doi.org/10.1016/j.jcp.2018.10.045} {\path{doi:https://doi.org/10.1016/j.jcp.2018.10.045}}.
\newline\urlprefix\url{https://www.sciencedirect.com/science/article/pii/S0021999118307125}

\bibitem{SHU2023111972}
D.~Shu, Z.~Li, A.~{Barati Farimani}, \href{https://www.sciencedirect.com/science/article/pii/S0021999123000670}{A physics-informed diffusion model for high-fidelity flow field reconstruction}, Journal of Computational Physics 478 (2023) 111972.
\newblock \href {https://doi.org/https://doi.org/10.1016/j.jcp.2023.111972} {\path{doi:https://doi.org/10.1016/j.jcp.2023.111972}}.
\newline\urlprefix\url{https://www.sciencedirect.com/science/article/pii/S0021999123000670}

\bibitem{PARISI1981378}
G.~Parisi, \href{https://www.sciencedirect.com/science/article/pii/0550321381900560}{Correlation functions and computer simulations}, Nuclear Physics B 180~(3) (1981) 378--384.
\newblock \href {https://doi.org/https://doi.org/10.1016/0550-3213(81)90056-0} {\path{doi:https://doi.org/10.1016/0550-3213(81)90056-0}}.
\newline\urlprefix\url{https://www.sciencedirect.com/science/article/pii/0550321381900560}

\bibitem{400cc48a-c98e-3a92-ac96-5477dc6f3a71}
U.~Grenander, M.~I. Miller, \href{http://www.jstor.org/stable/2346184}{Representations of knowledge in complex systems}, Journal of the Royal Statistical Society. Series B (Methodological) 56~(4) (1994) 549--603.
\newline\urlprefix\url{http://www.jstor.org/stable/2346184}

\bibitem{songblog}
Y.~Song, \href{https://yang-song.net/blog/2021/score/}{Estimating gradients of the data distribution} (05 2021).
\newline\urlprefix\url{https://yang-song.net/blog/2021/score/}

\bibitem{ANDERSON1982313}
B.~D. Anderson, \href{https://www.sciencedirect.com/science/article/pii/0304414982900515}{Reverse-time diffusion equation models}, Stochastic Processes and their Applications 12~(3) (1982) 313--326.
\newblock \href {https://doi.org/https://doi.org/10.1016/0304-4149(82)90051-5} {\path{doi:https://doi.org/10.1016/0304-4149(82)90051-5}}.
\newline\urlprefix\url{https://www.sciencedirect.com/science/article/pii/0304414982900515}

\bibitem{6795935}
P.~Vincent, A connection between score matching and denoising autoencoders, Neural Computation 23~(7) (2011) 1661--1674.
\newblock \href {https://doi.org/10.1162/NECO_a_00142} {\path{doi:10.1162/NECO_a_00142}}.

\bibitem{numericalmethodsSDEs}
P.~Kloeden, E.~Platen, Numerical Solution of Stochastic Differential Equations, Stochastic Modelling and Applied Probability, Springer Berlin, Heidelberg, 2011.

\bibitem{Zhu_2018}
Y.~Zhu, N.~Zabaras, {Bayesian Deep Convolutional Encoder–Decoder Networks for Surrogate Modeling and Uncertainty Quantification}, Journal of Computational Physics 366 (2018) 415–447.

\bibitem{karras2022elucidating}
T.~Karras, M.~Aittala, T.~Aila, S.~Laine, \href{https://openreview.net/forum?id=k7FuTOWMOc7}{Elucidating the design space of diffusion-based generative models}, in: A.~H. Oh, A.~Agarwal, D.~Belgrave, K.~Cho (Eds.), Advances in Neural Information Processing Systems, 2022.
\newline\urlprefix\url{https://openreview.net/forum?id=k7FuTOWMOc7}

\bibitem{esser2024scaling}
P.~Esser, S.~Kulal, A.~Blattmann, R.~Entezari, J.~M{\"u}ller, H.~Saini, Y.~Levi, D.~Lorenz, A.~Sauer, F.~Boesel, et~al., Scaling rectified flow transformers for high-resolution image synthesis, arXiv preprint arXiv:2403.03206 (2024).

\bibitem{takamoto2022pdebench}
M.~Takamoto, T.~Praditia, R.~Leiteritz, D.~MacKinlay, F.~Alesiani, D.~Pfl{\"u}ger, M.~Niepert, Pdebench: An extensive benchmark for scientific machine learning, Advances in Neural Information Processing Systems 35 (2022) 1596--1611.

\bibitem{ho2022classifier}
J.~Ho, T.~Salimans, Classifier-free diffusion guidance, arXiv preprint arXiv:2207.12598 (2022).

\bibitem{chaturantabut2010nonlinear}
S.~Chaturantabut, D.~C. Sorensen, Nonlinear model reduction via discrete empirical interpolation, SIAM Journal on Scientific Computing 32~(5) (2010) 2737--2764.

\bibitem{wentland2023scalable}
C.~R. Wentland, K.~Duraisamy, C.~Huang, Scalable projection-based reduced-order models for large multiscale fluid systems, AIAA Journal 61~(10) (2023) 4499--4523.

\end{thebibliography}
